\documentclass[10pt,journal,compsoc]{IEEEtran}
\ifCLASSOPTIONcompsoc
 \usepackage[nocompress]{cite}
\else
 \usepackage{cite}
\fi

\usepackage{adjustbox}
\usepackage{supertabular}
\usepackage{geometry}
\usepackage{array}
\usepackage{pifont}

\usepackage{multirow}
\usepackage[T1]{fontenc}
\usepackage{graphicx}
\usepackage[pagebackref=true,colorlinks,bookmarks=false,citecolor=blue,linkcolor=blue]{hyperref}

\ifCLASSINFOpdf
\else
\fi
\begin{document}
\title{Weakly Supervised Object Localization and Detection: A Survey}

\author{Dingwen Zhang,
 Junwei Han,
 Gong Cheng,
 %~\IEEEmembership{Senior Member,~IEEE}
 and Ming-Hsuan Yang %~\IEEEmembership{Fellow,~IEEE}% <-this % stops a space

\thanks{D. Zhang, J. Han, and G. Cheng are with Brain and Artificial Intelligence Laboratory, School of Automation, Northwestern Polytechnical University, Xi'an, China. Webpage: {https://nwpu-brainlab.gitee.io/index\_en.html}.}
\thanks{M.-H. Yang is with EECS, University of California at Merced, Merced, California United States 95344. E-mail: mhyang@ucmerced.edu.}
\thanks{This work is supported in part by the National Key R$\&$D Program of China under Grant 2017YFB1002201, the National Science Foundation of China under Grants 61876140 and 61773301. M.-H. Yang is supported in part by NSF CAREER Grant 1149783. (Corresponding author: Junwei Han.)}
}

\markboth{IEEE Transactions on Pattern Analysis and Machine Intelligence}%
{Shell \MakeLowercase{\textit{et al.}}: Bare Demo of IEEEtran.cls for Computer Society Journals}

\IEEEtitleabstractindextext{%
\begin{abstract}
As an emerging and challenging problem in the computer vision community, weakly supervised object localization and detection plays an important role for developing new generation computer vision systems and has received significant attention in the past decade.
As methods have been proposed, a comprehensive survey
of these topics is of great importance.
%
%This paper reviews the recent advances in the field of weakly supervised object localization and detection.
%
In this work, we review (1) classic models, (2) approaches with feature representations from off-the-shelf deep networks, (3) approaches solely based on deep learning, and (4) publicly available datasets and standard evaluation metrics that are widely used in this field.
We also discuss the key challenges in this field, development history of this field, advantages/disadvantages of the methods in each category, the relationships between methods in different categories, applications of the weakly supervised object localization and detection methods, and potential future directions to further promote the development of this research field.
%
%It is our hope that this survey will be beneficial for the researchers to have better understanding of this research field.
\end{abstract}

}

\maketitle

\IEEEdisplaynontitleabstractindextext

% \fi
\IEEEpeerreviewmaketitle

\IEEEraisesectionheading{\section{Introduction}\label{sec:introduction}}

\begin{figure}[t]
 \centering
 \includegraphics[width=1\linewidth]{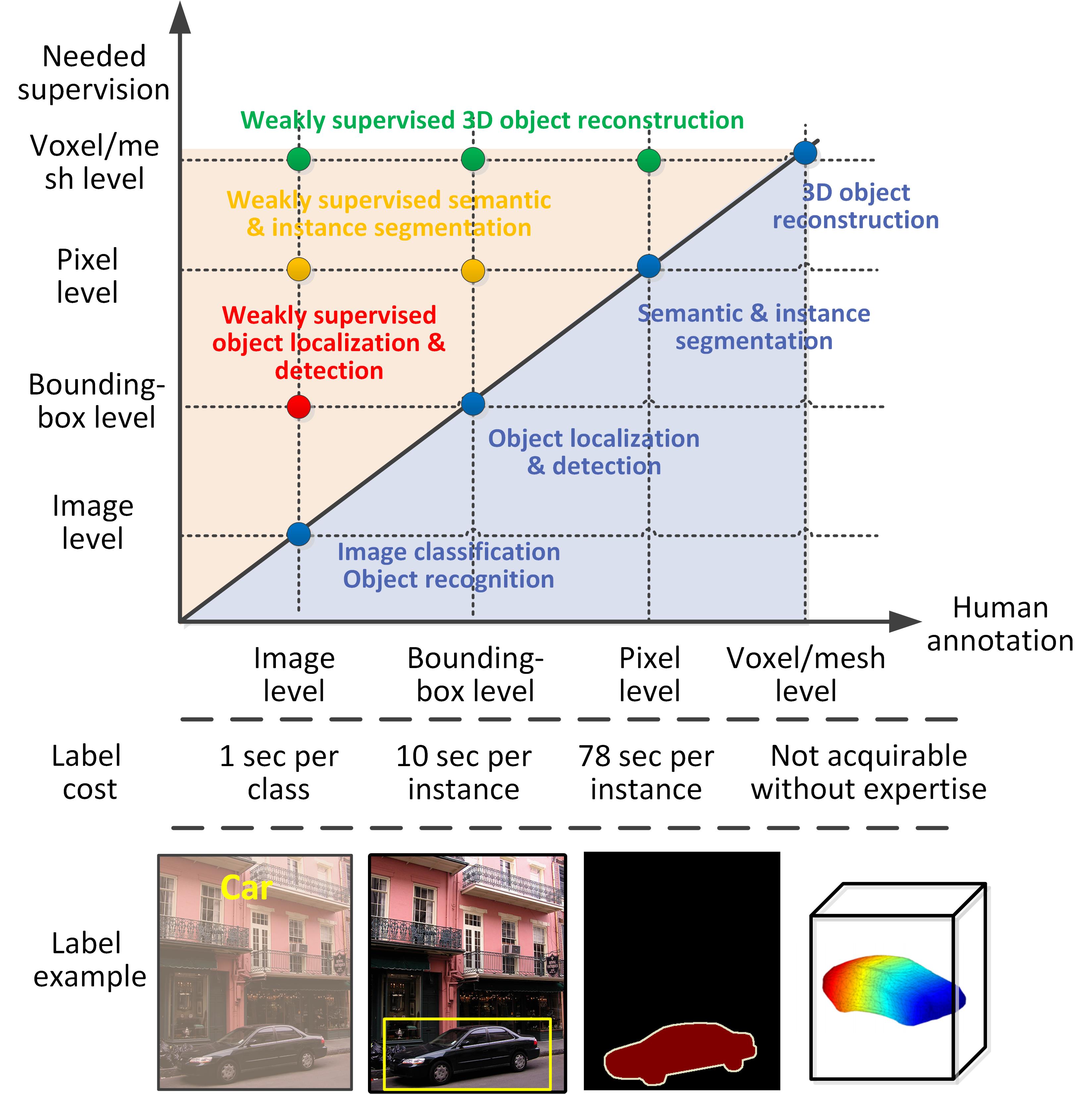}
 \caption{Illustration of the weakly supervised learning tasks in computer vision community. The blue area in the top block indicates the conventional fully-supervised learning tasks, while the red area in the top block indicates the weakly supervised learning tasks. The coordinate axes shows different levels of human annotation or supervision requirement, from low cost to high cost. Notice that the high cost annotation can be transformed to low cost annotation easily, e.g., from bounding-box level to image level, whereas the low cost annotation is hard to be transformed to high cost annotation. In the bottom block, we also show the label cost, in terms of annotation time, and the examples of different type of annotations. In this survey, we mainly focus on reviewing the research progress in weakly supervised object localization and detection, i.e., the red dot in the top block.}
 \label{illustration}
% \vspace{-0.6cm}
\end{figure}

\IEEEPARstart{W}{eakly} supervised learning (WSL) has recently received much attention in computer vision community.
A plethora of methods on this topic have been proposed in the past decade to address the challenging computer vision tasks including semantic segmentation~\cite{Chen2018DeepLab}, object detection~\cite{Ren2017Faster}, and 3D reconstruction~\cite{Fan2017A}, to name a few.
As shown in Fig.~\ref{illustration}, a WSL problem is defined as the learning process when some partial information regarding the task (e.g., class label or object location) on a small subset of the data points is at our disposal.
Compared to the conventional learning framework, e.g., fully supervised learning approaches, the WSL framework needs to operate on the small amount of weakly-labelled training data to learn the target model, which alleviates a huge amount of human labor to annotate training samples.
It can also facilitate the learning process when the fine-grained annotation
is extremely labor intensive and time consuming to even obtain the whole labeled data required by the fully-supervised approaches.

While a plethora of WSL-based vision methods have been developed, this survey mainly focuses on the task of weakly supervised object localization and detection, which is shown as the red dot in Fig.~\ref{illustration}.
It is well-known that object localization and detection is a fundamental research problem in computer vision.
Learning object localization and detection models under weak supervision has attracted much attention in the past decades.
%a widely-studied direction
%in the research field of object localization and detection and it was also one of the first WSL-based computer vision task with early methods dating back to 17 years ago.
%
While existing methods treat weakly supervised object localization (WSOL) and weakly supervised object detection (WSOD) as two different tasks\footnote{ {The difference between WSOL and WSOD mainly lies in that WSOL mainly aims at localizing a single known (or unknown) object from each given image scene. The goal of WSOD is to instead detect every possible object instance from the given image scene. This makes WSOD a little more difficult than WSOL.}}, we consider these as a common task due to several reasons: 1) these tasks learn with the same image-level human annotation; 2) these two tasks need certain supervision as input and usually aim to localize objects on the bounding-box level as output; 3) WSOD task can be accomplished by directly training off-the-shelf fully supervised object detectors on the object locations obtained from WSOL.

During the last decade, considerable efforts have been made to develop various approaches for learning object detectors with weak supervision.
Some of the existing algorithms only learn weakly supervised object detectors for one or several certain object categories, such as vehicles~\cite{cao2016weakly}, traffic signs~\cite{krapac15,zadrija2015patch}, pedestrians~\cite{wang2013weakly,blaschko2010simultaneous}, faces~\cite{hoai2014learning,galleguillos2008weakly}, tuberculosis bacilli~\cite{hwang2016self}, aircrafts~\cite{han2015object,Zhang2014weakly,zhou2015negative,zhou2016weakly}, and human actions~\cite{mathe2014,hoai2014learning}.
While other approaches, e.g., \cite{bilen2016weakly,ren2016weakly,wang2015large}, focus on developing weakly supervised learning frameworks for unconstrained object categories, i.e., learning frameworks can be extended to learn object detector for the given category-specific weakly-labelled training images.
As enormous methods have been developed for these important tasks, a comprehensive review of the literature concerning weakly supervised object localization and detection is of great importance.
%
%To this end, this paper aims to provide a survey of the research progress in this field for the past decades.

As weakly supervised object localization and detection methods mainly exploit the image-level manual annotation, the learning frameworks not only need to address the typical issues, such as the intra-class variations in appearance, transformation, scale and aspect ratio, encountered in conventional fully supervised object localization and detection task, but also the \textbf{learning under uncertainty} challenges caused by the inconsistency between human annotations and real supervisory signals.
In weakly supervised object localization and detection, the accuracy of object locations and learning processes are closely related.
The key is to propagate the image-level supervisory signals to the instance-level (bounding-box-level) training data for the learning processes.
As each training image can be labeled by numerous bounding boxes of different accuracy, propagating such weak supervision inevitably involves a large amount of ambiguous and noisy information as each training instance.
More specifically, the \textbf{learning under uncertainty} issue would cause the following challenges that make the weakly supervised learning process challenging:

\begin{itemize}
 \item \textbf{Learning with inaccurate instance locations:} This issue is mainly caused by the definition ambiguity in object parts and context.
 Without precise annotation or definition, it is difficult for a learner to decide whether an object category label associates with a discriminative object part, the whole object region, or the object with a certain context region.
 As a result, the bounding-box instance locations inferred by the learner may contain many inaccurate samples including the ones with local object parts or undesired contextual regions.
 These samples would negatively affect the performance of WSL-based detectors.
 \item \textbf{Learning with noisy samples:} Even when the bounding-box locations can be precisely labeled, the training examples enclosed by bounding-boxes may still be noisy as background pixels are usually included.
 As there is no additional information to separate foreground objects from the background, the learner may tag a ``background'' label to an object region when it fails to recognize the object category.
 In addition, the learner may mistakenly label a bounding-box that contains a bicycle as a motorcycle, as these two object categories share many similar features.
 %
 %Such noisy label set would also influence the learning process of the object detector.
 %
 \item \textbf{Learning with domain shifts:} For a certain object category, the image regions localized during the learning process may only contain samples with limited diversity in object shape, appearance, scale, and view angle.
 This makes the subsequent learning process biased to limited knowledge of the object category and does not generalize well for test samples.
 For instance, a learner can hardly localize or detect a flying swan when all the training samples contain the swimming ones on lakes.
 This issue happens frequently among the weakly supervised learning process when there is a large gap between the training and testing domains.
 \item \textbf{Learning with insufficient instance samples:}
 %From the perspective of the training samples, learning object detectors under weak supervision could only rely on image-level samples or instance-level samples inferred from the image.
 %
 %Notice that it is difficult to successfully extract all the instance-level samples from a certain image if the image contains complex background and multiple interactive object instances.
 %
 %As a result, the samples, especially the ones with accurate bounding-box locations and labels, used to learn weakly supervised object detectors would be much less than those used in learning the conventional fully supervised object detectors.
 %
 Similar to the issues in conventional learning methods, it is difficult to train effective object detectors under the weakly-supervised setting when the amount of training samples is limited.
 In addition, the number of positive samples is usually much smaller than that of negative samples for binary classes.
 Furthermore, the data distributions for a large number of categories is usually long-tailed.
 This issues are significantly exaggerated for the WSL-based methods using deep learning.
 %
 %This would also influence the final learning performance especially when the weakly supervised learning framework is built based on the data-hunger learning models such as deep neural network.
 %
 %MH: I do not quite understand what you want to say here. If the instances are not detected in the first round, how will you use the images for the subsequent learrning?
 %DW: Here I want to say that some instances would be missed (remain unlabelled) in early round. So the refinement process need have the mechanism to gradually label these data and involve them in learning.
 %\item \textbf{Learning with unlabelled instance samples:} As the instance samples cannot be localized or detected all at once, there will be many instance samples left unlabelled during learning. Thus, the weakly supervised learning process should have the capacity to gradually recognize the previously unlabelled instance samples and involve them into the subsequent training phase. From this perspective, the weakly supervised learning process should be able to address the challenges in semi-supervised learning~\cite{tang2018visual}. Besides, considering the long tail problem between different object categories~\cite{ouyang2016factors}, the learner may recognize a relative larger number of training instances for some object categories while only a very small number of training instance for some other categories. This also causes the challenges in one/few shot learning~\cite{metaGAN2018,vinyals2016matching}.
\end{itemize}

To address the above-mentioned issues in learning weakly supervised object detectors, existing methods are usually constructed based on two steps: initialization and refinement.
The initialization stage is used to leverage certain prior knowledge to propagate image-level annotation into instance level, and thus can generate instance-level annotation (but with label noise, sample bias, and limited quality in location accuracy) for the learning process.
The refinement stage is used to leverage new instance samples obtained from the first stage to mine truthful knowledge about the objects of interest gradually and finally obtain the desired object models for localization and detection.
These two learning stages need to collaborate to address the aforementioned five-fold challenges.
In initialization stage, efforts should be made to improve the annotation quality as much as possible to generate training instances with proper locations, accurate labels, high diversity, and high recall rate.
As the annotation quality obtained in the learning stage cannot be perfect, in the refinement stage, further efforts should be made to improve the learner's robustness to cope with the inaccurate instance location, noisy examples, biased instance sample, insufficient instance sample issues as well as the capacity to take advantage of the unlabelled instance samples.
When properly addressing the problems in each learning stage,
good weakly supervised object detectors can be learned.

In this work, we review the existing weakly supervised object localization and detection approaches\footnote{Some early methods, such as~\cite{fergus2007weakly,crandall2006weakly}, learn to localize category-wise key points under the weak supervision, while this survey mainly focuses on the methods for localizing instances with bounding-boxes.}, which are divided into three main categories and eight subcategories.
These three main categories are based on classic approaches, feature representations from off-the-shelf deep models, and deep learning frameworks.
The eight subcategories include approaches for initialization, refinement, initialization and refinement, pre-trained deep features, inherent cues in deep models, fine-tuned deep models, single-network training, and multi-network training.
We further discuss the relationship between the approaches in different categories.
In addition, we also discuss open problems and challenges of current studies and propose several promising research directions in the future for constructing more effective weakly supervised object localization and detection frameworks.
%
%To the best of our knowledge, this is the first survey paper in the literature that focuses on weakly supervised object localization and detection.

\begin{figure*}[t]
 \centering
 \includegraphics[width=1\linewidth]{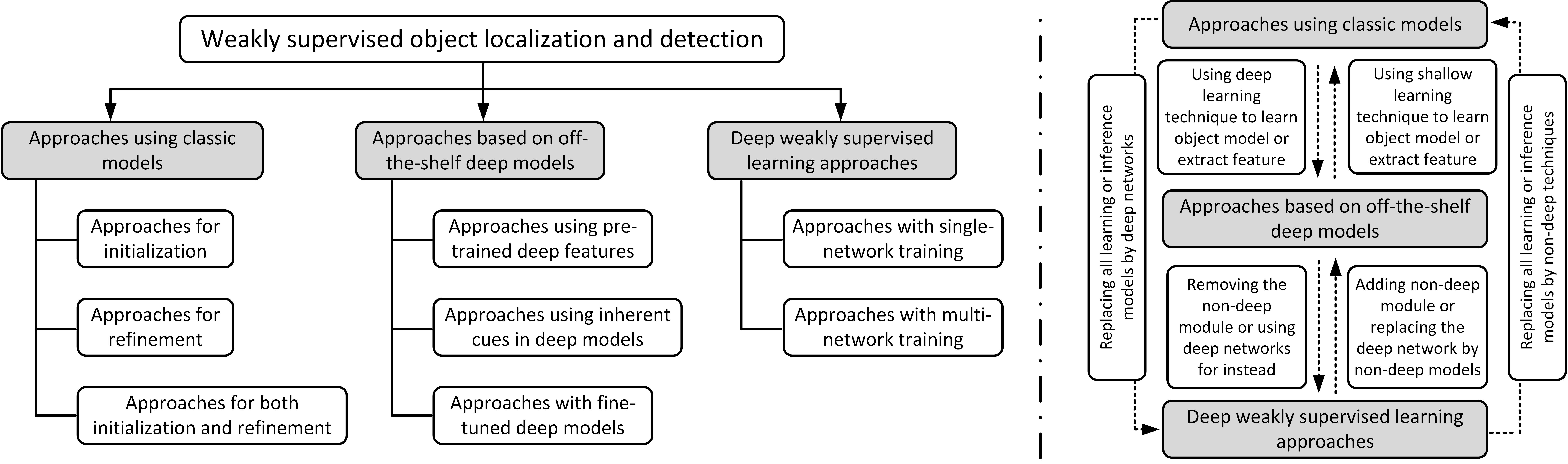}
 \caption{In the left block, taxonomy of the existing approaches for weakly supervised object localization and detection, which includes three main
 categories and eight subcategories.
 In the right block, the relationships between the approaches in different categories are shown.
}
 \label{taxonomy}
% \vspace{-0.6cm}
\end{figure*}

%The rest of the paper is organized as follows. Section 2 briefly introduces the taxonomy of methods for weakly supervised object localization and detection, together with the relationship between different categories. In Sections 3 - 5, we exhaustively review approaches solely based on non-deep learning technique, approaches using both deep and non-deep learning technique, and approaches solely based on deep learning technique, respectively. In Section 6, we review publicly available datasets and standard evaluation metrics that are widely used in this research area. Section 7 discusses the challenges of current studies and proposes several promising research directions to advance the field. Finally, conclusions are drawn in Section 9.
%
%
\begin{figure}[t]
 \centering
 \includegraphics[width=1\linewidth]{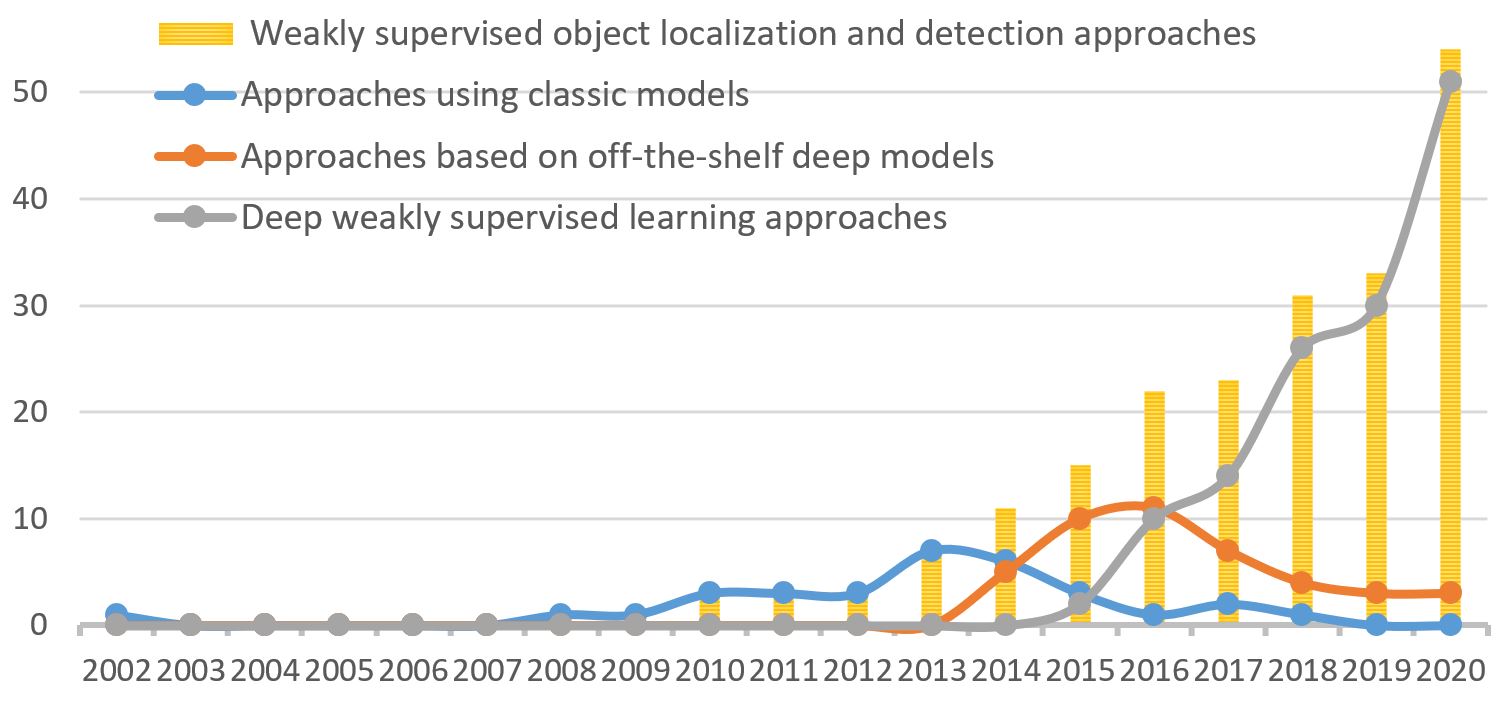}
 \caption{Developments of weakly supervised localization and detection methods. The yellow histogram shows the number of publications in this research field in each year, and the curves show the number of proposed methods each year for a particular category of approach.}

 \label{history}
% \vspace{-0.6cm}
\end{figure}

\section{Taxonomy}
\label{Taxonomy}
In the last decade, a plethora of methods have been developed for weakly supervised object localization and detection.
We can generally categorize existing methods based on classic formulations, feature representations from off-the-shelf deep models, and deep weakly supervised learning algorithms.
While inside each main category, we further divide the approaches into two or three subcategories.
Fig.~\ref{taxonomy} shows our taxonomy of the studies in the research field of weakly supervised object localization and detection.
In addition, Fig. \ref{history} reviews the development history of each of the main category as well as that of the whole research field.
A few approaches based on classic formulations appeared around 2002.
From 2002 to 2009, the research in this field went through a very slow pace.
Since 2014, numerous approaches based on both classic formulations and learned feature representations from deep models have been developed and received much attention.
While in the last few years, more approaches solely based on deep learning have become the main stream to address the problems of weakly supervised object localization and detection.
%
%Although numerous approaches have been presented in the past decades, the overall development of this research field still gains increasing attention.
While a plethora of methods have been developed to address different aspects of these problems in the past decades, this field is gaining increasing attention.

\renewcommand{\multirowsetup}{\centering}
 \begin{table*} [t]

 \tiny
 \caption{Summary of the approaches for initialization, which is a subcategory in the weakly supervised object localization and detection approaches that learn by classic models. An approach is considered for general object category when it is tested for detecting more than five object categories in the corresponding literature. The approaches with None detector indicate the weakly supervised object localization approaches.}
 \label{table initialization}
 \resizebox{\linewidth}{!}{\begin{tabular}{m{2.5cm}<{\centering}m{1.2cm}<{\centering}m{1.5cm}<{\centering}m{3.1cm}<{\centering}m{1.8cm}<{\centering}m{1.5cm}<{\centering}m{3.5cm}<{\centering}m{1.8cm}<{\centering}}
 \hline
 \multirow{2}{*}{\textbf{Methods}} & \multirow{2}{*}{\textbf{Detector}} & \multirow{2}{*}{\textbf{Descriptor}} & \multirow{2}{*}{\textbf{Prior knowledge}} & \multirow{2}{*}{\textbf{Extra training data}} & \multirow{2}{*}{\textbf{Learning model}} & \multirow{2}{*}{\textbf{Learning strategy}} & \multirow{2}{*}{\textbf{Object category}} \\
 & & & & & & &\\
\hline \hline
Cao-PR-2017\cite{cao2017weakly} & SVM & HOG+PCA & Road map prior + density prior & None & SVM & MIL with density estimation & Vehicle in satellite imagery \\
%\hline
 %\hline
Shi-TPAMI-2015\cite{shi2015bayesian} & DPM & SIFT+Lab+LBP, BOW & Appearance prior + geometry prior & None & Topic model & Bayesian inference & General objects \\
%\hline
Tang-ICIP-2014\cite{tang2014fusing} & DPM & HOG & Saliency (objectness score) & None & DPM & Select initial boxes + DPM training & General objects \\
%\hline
 Xie-VCIP-2013\cite{xie2013object} & None & SIFT & Intra-class consistency & None & None & Low-rank and sparse coding & General objects \\
 %\hline
Siva-CVPR-2013\cite{siva2013looking} & DPM & Lab+SIFT & Saliency & Labelme + PASCAL07, 12 (unlabelled) &None & None & General objects \\
%\hline
 Shi-ICCV-2013\cite{shi2013bayesian} & DPM & SIFT & Appearance prior (spatial distribution, size, saliency) & None & Topic model & Bayesian inference & General objects \\
%\hline
 Sikka-FG-2013\cite{sikka2013weakly} & None & SIFT, LLC, BOW & None & None & MilBoost & Generating multiple segments for initialization and use Milboost for learning & Pain (on face) \\
%\hline
 Siva-ECCV-2012\cite{siva2012defence}& None & SIFT+BOW & Iter-class variance + saliency & None & None & Negative mining & General objects \\
 Shi-BMVC-2012 \cite{shi2012transfer}& None & BOW & Mapping relationship between the box overlap and appearance similarity & Part of PASCAL 07 (box annotation) & RankSVM & Transfer learning by ranking & General objects \\
%\hline
 Khan-AAPRW-2011\cite{khan2011learning} & DPM & Phog/phow & None & Internet image (weakly annotated) & MIL & Learning from internet image & Pascal@8 \\
%\hline
Zhang-BMVC-2010\cite{zhang2010weakly} & SVM & IHOF & Co-occurrence & None & SVM & High order feature learning by exploring co-occurrence & General objects \\
\hline
 \end{tabular}}
\end{table*}

\renewcommand{\multirowsetup}{\centering}
\begin{table*} [t]
 \tiny
	\caption{Summary of the approaches for refinement, which is a subcategory in the weakly supervised object localization and detection approaches that learn by classic models.
	Here, * indicates a certain variation of the corresponding model. An approach is considered for general object category when it is tested for detecting more than five object categories in the corresponding literature. The approaches with None detector indicate the weakly supervised object localization approaches.}
	 \label{table refinement} \resizebox{\linewidth}{!}{\begin{tabular}{m{2.5cm}<{\centering}m{1.5cm}<{\centering}m{3cm}<{\centering}m{2cm}<{\centering}m{2cm}<{\centering}m{2cm}<{\centering}m{3.1cm}<{\centering}m{2.5cm}<{\centering}}
\hline
 \multirow{2}{*}{\textbf{Methods}} &
 \multirow{2}{*}{\textbf{Detector}} &
 \multirow{2}{*}{\textbf{Descriptor}} &
 \multirow{2}{*}{\textbf{Prior knowledge} } &
 \multirow{2}{*}{\textbf{Extra training data}} &
 \multirow{2}{*}{\textbf{Learning model}} &
 \multirow{2}{*}{\textbf{Learning strategy}} &
 \multirow{2}{*}{\textbf{Object category}} \\
 & & & & & & &\\
\hline
\hline
 Wang-TMI-2018\cite{wang2018weakly} & None & Color & None & None & Low-rank model & Low-rank Factorization & General lesion \\
%\hline
 Wang-TC-2017\cite{wang2017instance} & None & SIFT+LAB & None & None & Probability model & BOW learning+instance labeling & General objects \\
%\hline
 Cholakkal-CVPR-2016\cite{cholakkal2016backtracking} & None & SIFT & Saliency & None & SVM* & ScSPM-based top down saliency & Salient objects \\
%\hline
 Zadrija-GCPR-2015\cite{zadrija2015patch} & None & SIFT+FV & None & None & GMM + linear classifier & Patch-level spatial layout learning & Traffic sign \\
%\hline
 Krapac-ICCVTA-2015\cite{krapac2015weakly} & None & SIFT+FV & None & None & Sparse logistic regression & Sparse classification & Traffic sign \\
%\hline
 Cinbis-CVPR-2014\cite{gokberk2014multi} & SVM & FV & Center prior & None & SVM & Multi-fold MIL & General objects \\
%\hline
 Wang-ICIP-2014\cite{wang2014weakly} & SVM + graph model & SIFT & None & None & SVM + graph model & Maximal entropy random walk & Car, dog \\
%\hline
 Wang-ICIP-2014\cite{wang2014window} & SVM & SIFT & None & None & SVM & Clustering for window mining & General objects \\
%\hline
 Tang-CVPR-2014\cite{tang2014co} & None & SIFT & Saliency & None & Boolean constrained quadratic program & Mine similarity and discriminativeness both for image and box & General objects \\
 Hoai-PR-2014\cite{hoai2014learning} & SVM & SIFT,BOW & None & None & SVM* & Localization-classification SVM & Face,car, human motion \\
%\hline
 Wang-WACV-2013\cite{wang2013weaklyWacv} & Task-specific detectors & HOG/SC & Background\ saliency & None & MIL+AdaBoost* & Soft-label Boosting after MIL & Vehicle, pedestrain\\
%\hline
%\hline
 Kanezaki-MM-2013\cite{kanezaki2013weakly} & Linear classifiers & 3D voxel feature (color+C3HLAC +Intensity, texture, GRSD) & None & None & Linear classifiers & Multi-class MIL & Balls, tools \\
%\hline
%\hline
 Pandey-ICCV-2011\cite{pandey2011scene} & DPM* & HOG & None & None & DPM* & Learning DPM with fully latent variable & General objects \\
%\hline
Blaschko-NIPS-2010\cite{blaschko2010simultaneous} & None & BOW/HOG & None & None & SVM* & Learning SVM with structured output ranking objective & Cat, pedestrian \\
%\hline
%\hline
Hoai-ICCV-2009\cite{nguyen2009weakly} & SVM & SIFT,BOW & None & None & SVM* & Localization-classification SVM & Face,car, human motion \\
%\hline
 Galleguillos-ECCV-2008\cite{galleguillos2008weakly} & None & SIFT+BOW & None & None & MilBoost & Train MIL classifier for localization & Landmarks, faces, airplanes, leopard, motorbike, car \\
%\hline
 Rosenberg-BMVC-2002 & GMM & Orientation deriviate filters & Exampler prior & Training exampler (box annotation) & GMM & Learning from exampler training data to weakly labelled training data & Telephone \\
\hline
 \end{tabular}}
\end{table*}
\renewcommand{\multirowsetup}{\centering}
\begin{table*} [t]
 \tiny
	\caption{Approaches for both initialization and refinement, which is a subcategory in the weakly supervised object localization and detection approaches that learn by classic models. An approach is considered for general object category when it is tested for detecting more than five object categories in the corresponding literature. The approaches with None detector indicate the weakly supervised object localization approaches.}
	 \resizebox{\linewidth}{!}{\begin{tabular}{m{1.8cm}<{\centering}m{1.2cm}<{\centering}m{1.5cm}<{\centering}m{2cm}<{\centering}m{1.8cm}<{\centering}m{1.5cm}<{\centering}m{3cm}<{\centering}m{1.5cm}<{\centering}}
\hline
 \multirow{2}{*}{\textbf{Methods}} &
 \multirow{2}{*}{\textbf{Detector}} &
 \multirow{2}{*}{\textbf{Descriptor}} &
 \multirow{2}{*}{\textbf{Prior knowledge} } &
 \multirow{2}{*}{\textbf{Extra training data}} &
 \multirow{2}{*}{\textbf{Learning model}} &
 \multirow{2}{*}{\textbf{Learning strategy}} &
 \multirow{2}{*}{\textbf{Object category}} \\
 & & & & & & &\\
\hline
\hline
 Wang-cvpr-2013\cite{wang2013weakly} & None & Color+SIFT & None & None & HST+SVM & Joint parsing and attribute localization & Scene attributes \\
%\hline
 Deselaers-IJCV-2012\cite{deselaers2012weakly} & DPM & GIST+CH+BOW+HOG & Generic knowledge & Meta-training data with box annotation & CRF+DPM & Learning appearance model by trainsferring genearic knowledge & General objects \\
 Siva-ICCV-2011\cite{siva2011weakly} & DPM & SIFT+BOW+HOG & Inter-class prior + intra-class prior & None & DPM & Model drift learning & General objects \\
Deselaers-ECCV-2010\cite{deselaers2010localizing} & DPM & GIST+CH+BOW+HOG & Generic knowledge & Meta-training data with box annotation & CRF+DPM & Learning appearance model by transferring generic knowledge & General objects \\ \hline
 \end{tabular}}
\end{table*}

%\subsection{Relationship Between Main Categories}

As shown in the right block of Fig.~\ref{taxonomy}, methods in
main categories are related in several aspects.
%
%For the approaches learn by non-deep techniques, if we replace a component from their frameworks, such as the learning model or feature extractor, by a deep neural network, we can then obtain the approaches learn by both deep network and non-deep techniques.
%
%Similarly, for the approaches learn by both deep network and non-deep techniques, if we replace the component implemented based on deep networks by the non-deep techniques, these approaches would become to the approaches learn by non-deep techniques.
%
Numerous methods are developed based on the classic formulations with the advances of feature representations from deep models.
%
%For the relation between the second category and the third category, if we remove the non-deep modules from the approaches in the second category, i.e., the approaches learn by both deep network and non-deep techniques, or using deep networks for instead, we would obtain the approaches in the third category, i.e., the approaches solely based on deep learning. If we add non-deep modules or replacing the deep network in the approaches solely based on deep learning by non-deep models, the approaches originally belonging to the third category would become to the approaches belonging to the second category.
Similarly, a number of methods based solely on deep models are end-to-end trainable by considering classic formulations and feature extraction schemes.

%For relation between the first category and the third category, if all learning or inference models in the approaches learn by non-deep techniques are replaced by deep networks, then the approaches originally belonging to the first category would become to the approaches belonging to the third category. Similarly, if all learning or inference models in the approaches solely based on deep learning are replaced by non-deep techniques, then the approaches originally belonging to the third category would become to the approaches belonging to the first category.

%From the above discussion, we can observe that all the three categories of approaches can be converted to each other with certain conditions.
%
%Besides, although the first-category approach, i.e., the approaches learn by non-deep techniques, and the third-category approach, the approaches solely based on deep learning, can be converted to each other directly, the second-category approach, the approaches learn by both deep network and non-deep techniques, can also be considered as an intermediate station to facilitate such conversion. This actually in line with the order of the appearance of different weakly supervised object location and detection methods in the development history of this field (see Fig. \ref{history}).

\begin{figure}[t]
 \centering
 \includegraphics[width=7 cm]{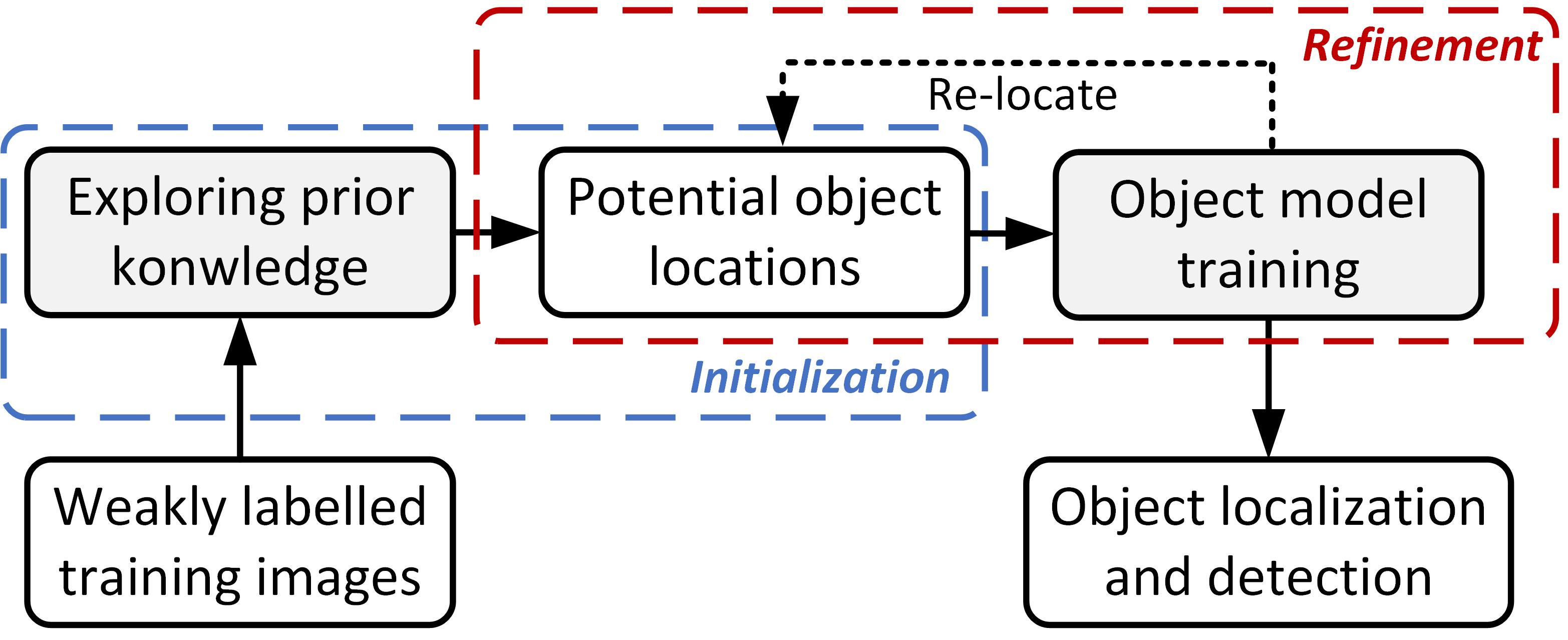}
 \caption{Flowchart of the weakly supervised object localization and detection approaches using classic models.}
 \label{nondeep flowchart}
 %\vspace{-0.6cm}
\end{figure}

\section{Classic Models}
\label{Classic Models}
In this section, we review the classic approaches that learn weakly supervised object localizer or detector without using deep features.
These methods typically consist of one initialization module followed by one refinement process as shown in
Fig.~\ref{nondeep flowchart}.
In \cite{siva2013looking,shi2013bayesian,khan2011learning,deselaers2012weakly,deselaers2010localizing}, the detector is based on the deformable part model (DPM)~\cite{felzenszwalb2010object}.
In other approaches \cite{cao2017weakly,zhang2010weakly,gokberk2014multi,wang2014weakly}, the detector is based on the support vector machine (SVM) classifier.
The features used by these approaches are hand-crafted feature descriptors, such as HOG in \cite{cao2017weakly,tang2014fusing,zhang2010weakly,wang2013weaklyWacv}, SIFT in \cite{siva2013looking,sikka2013weakly,siva2012defence,wang2017instance}, and Lab color in \cite{wang2017instance,siva2013looking,shi2015bayesian}, which are sometimes used to build higher level representation such as bag-of-words (BOW) in \cite{shi2015bayesian}, \cite{sikka2013weakly},\cite{shi2017transfer},\cite{siva2012defence},\cite{hoai2014learning}, Fisher vector representation in \cite{gokberk2014multi}, and subspace-based representation in \cite{cao2017weakly}.
In the following, we divide these approaches for initialization and refinement process.
%We will introduce each of the subcategory of approach in below.

\subsection{Initialization}
Numerous methods have been developed to mine reliable instance samples, using prior knowledge, as weak supervision for the following processes.
%
%The crux of these approaches is to explore useful prior knowledge for inferring potential objects of interest from the weakly labelled training images.
%
A brief summary of these approaches are shown in Table \ref{table initialization}.

Zhang et al.~\cite{zhang2010weakly} leverage the prior-knowledge of object co-occurrence to identify translation and scale invariant high order features for weakly supervised object localization.
In~\cite{shi2012transfer} Shi et al. propose a transfer learning paradigm to first use a RankSVM to learn the mapping relationship between the box overlap and appearance similarity from an auxiliary training data (with bounding-box level annotation) and then transfer the learned prior-knowledge for localizing objects of interest in the given weakly labeled images.
A simple yet effective approach, named as negative mining, is developed by Siva et al. to explore the inter-class variance among the object regions in weakly labeled training images.
The final object locations are obtained by
using a linear combination of the inter-class variance and saliency prior.
Similarly, Tang et al.~\cite{tang2014fusing} and Xie et al. \cite{xie2013object} use the saliency prior and intra-class consistency to mine the initial object locations, respectively.
In \cite{shi2013bayesian,shi2015bayesian}, Shi et al. explore the appearance prior and geometry prior in their topic model to build a Bayesian joint modeling framework for weakly supervised object localization.
On the other hand, Cao et al.~\cite{cao2017weakly} exploit the road map prior and density prior to mine the initial vehicle locations from the weakly labeled satellite images and then trained the vehicle detector under a modified multiple-instance learning (MIL) model.

\renewcommand{\multirowsetup}{\centering}
\begin{table*} [t]
 \tiny
	\caption{Summary of the approaches using pre-trained feature representations, which is a subcategory in the weakly supervised object localization and detection approaches based on the off-the-shelf deep models. * indicates a certain variation of the corresponding model. An approach is considered for general object category when it is tested for detecting more than five object categories in the corresponding literature. The approaches with None detector indicate the weakly supervised object localization approaches.}
 \label{table off-the-shelf feature}
 \resizebox{\linewidth}{!}{\begin{tabular}{m{2cm}<{\centering}m{1cm}<{\centering}m{1.6cm}<{\centering}m{2.8cm}<{\centering}m{3.8cm}<{\centering}m{1.2cm}<{\centering}m{3.5cm}<{\centering}m{2.2cm}<{\centering}}
\hline
 \multirow{2}{*}{\textbf{Methods}} &
 \multirow{2}{*}{\textbf{Detector}} &
 \multirow{2}{*}{\textbf{Descriptor}} &
 \multirow{2}{*}{\textbf{Prior knowledge} } &
 \multirow{2}{*}{\textbf{Extra training data}} &
 \multirow{2}{*}{\textbf{Learning model}} &
 \multirow{2}{*}{\textbf{Learning strategy}} &
 \multirow{2}{*}{\textbf{Object category}} \\
 & & & & & & &\\
\hline
\hline
 Gonthier-arxiv-2018\cite{gonthier2018weakly} & None & CNN & Supervised objectness score (Fast R-CNN(Resnet)) & ImageNet(tag label) & SVM & MIL & Objects in art (watercolor2K, people-art) \\
%\hline
Zadrija-CVIU-2018\cite{zadrija2018sparse} & None & VGG19 Conv 5\_4. SIFT, Fisher vector & None & ImageNet(tag label) & Sparse model & None & Zebra crossings, traffic signs \\
%\hline
 Cinbis-TPAMI-2017\cite{cinbis2017weakly} & SVM* & FV+CNN & Center prior & ImageNet(tag label) & SVM* & Multi-fold MIL & General objects \\
%\hline
Wei-IJCAI-2017\cite{wei2017deep} & None & CNN & None & ImageNet(tag label) & None & Deep descriptor transforming & General objects \\
%\hline
 Zhang-IJCAI-2016\cite{zhang2017bridging} & SVM & CNN & Saliency prior & ImageNet(tag label) & SVM & Easy-to-hard(SPL+CL) & General objects \\
%\hline
 Li-ECCV-2016 & None & FC6 & Strong detector prior (sparsity) & ImageNet(tag label) & SVM & Regularizing score distribution & General objects \\
%\hline
 Ren-TPAMI-2016\cite{ren2016weakly} & SVM* & FC6 & None & ImageNet(tag label) & SVM* & MIL+bag splitting & General objects \\
%\hline
 Wan-ICIP-2016\cite{wan2016weakly} & SVM* & FC7 & None & ImageNet(tag label) & SVM* & Correlation suppression+part suppression & General objects \\
%\hline
 Rochan-IVC-2016\cite{rochan2016weakly} & None & Color histogram +CNN & Saliency & PASCAL (edge box), ImageNet & SVM & None & General objects \\
%\hline
 Shi-ECCV-2016\cite{shi2016weakly} & SVM & FC7(Alexnet) & Size prior & ImageNet(tag label), PASCAL2012 (object size) & SVM & Easy-to-hard(curriculum) & General objects \\
%\hline
 Wang-TIP-2015\cite{wang2015large,wang2014weakly} & SVM & FC6 & None & ImageNet(tag label) & pLSA, SVM & Online latent category learning & General objects \\
%\hline
Rochan-CVPR-2015\cite{rochan2015weakly} & SVM & CNN & Objectness score, word embedding prior & YouTube-Objects (for parameter validation), Familiar object categories(detector) & SVM, Sparse reconstruction & Appearance transfer from text representation & General objects \\
%\hline
Bilen-CVPR-2015\cite{bilen2015weakly} & LSVM & FC7+spatial features & None & ImageNet(tag label) & LSVM & Convex clustering & General objects \\
%\hline
Wang-ICCV-2015\cite{wang2015relaxed} & None & FC6 & None & PASCAL(edge box), ImageNet & SVM* & Relaxed multiple-Instance SVM & General objects \\
%\hline
Zhou-ICMBD-2015\cite{zhou2015negative} & SVM & FC7 & Saliency prior & ImageNet(tag label) & SVM & Negative Bootstrapping & Airplanes in remote sensing \\
%\hline
Han-TGRS-2015\cite{han2015object} & SVM & DBM & Saliency, intra-class compactness, inter-class separability & None & DBM + SVM & Bayesian framework for initialization + refinement detector training & Objects in remote sensing \\
%\hline
Mathe-Arxiv-2014\cite{mathe2014multiple} & Sequential detector & FC6 & Human fixation & ImageNet(tag label) & MIL*+RL & Constrained multiple instance SVM learning + reinforcement learning of detector & Human actions \\
%\hline
 Wang-ECCV-2014\cite{wang2014weakly} & SVM & FC6 & None & ImageNet(tag label) & pLSA, SVM & Online latent category learning & General objects \\
%\hline
 Bilen-BMVC-2014\cite{bilen2014weakly} & LSVM & DeCAF & None & ImageNet(tag label) & LSVM* & LSVM with posterior regularization on symmetry and mutual exclusion & General objects \\
Song-NIPS-2014\cite{song2014weakly} & DPM & FC7 & Objectness score & ImageNet(tag label) & LSVM & Frequent configuration mining+detector training & General objects \\
%\hline
 Song-ICML-2014\cite{song2014learning} & LSVM & DeCAF & None & ImageNet(tag label) & Graph model+LSVM* & Initialization via discriminative submodular cover+smoothed LSVM learning & General objects \\
\hline
 \end{tabular}}
\end{table*}

\subsection{Refinement}
%These methods improve the refinement process for updating the object detectors and the corresponding object locations, while using the existing or naive techniques in the initialization stage.
After potential object instances are obtained, these hypotheses are verified in the following refinement processes.
The goals of these approaches are to design learning objective functions, optimization strategies, or learning mechanisms to gradually determine objects of interest from the extracted initial instance training samples.
A brief summary of these approaches is shown in Table \ref{table refinement}.

 Hoai et al. \cite{hoai2014learning,nguyen2009weakly} propose an approach which localizes the instances of the positive class and learns a sub-window classifier to recognize the corresponding object class.
Blaschko et al. \cite{blaschko2010simultaneous} use a structured output SVM to learn a regressor from the weakly labeled training images to object locations that are parameterized by the coordinates of the bounding boxes.
The object locations were treated as latent variables, while the image-level annotation was used to constrain the set of values the latent variable can take.
Similarly, Pandey et al. \cite{pandey2011scene} learn weakly supervised object detectors by using DPMs with latent SVM training.
In \cite{wang2013weaklyWacv}, a soft-label boosting approach is developed to exploit the soft labels that are estimated during the MIL process to train object detectors based on Boosting algorithm.
In \cite{tang2014co}, Tang et al. treat the weakly supervised object localization problem as an object co-localization task, and present a joint image-box formulation to mine reliable object locations via a Boolean constrained quadratic program.
This approach can handle noisy labels in the image-level annotations.
To address the property that the MIL process may converge to poor local optima after the initialization, Cinbis et al. \cite{gokberk2014multi} design a multi-fold MIL training paradigm.
This method divides the whole weakly labelled training images into multiple folds and implements the detector training process and object re-localization process in different folds, thereby alleviating the issue with convergence of poor local optima.

\renewcommand{\multirowsetup}{\centering}
\begin{table*} [t]
 \tiny
	\caption{Summary of the approaches using visual cues, which is a subcategory in the weakly supervised object localization and detection approaches based on the off-the-shelf deep models. * indicates a certain variation of the corresponding model. An approach is considered for general object category when it is tested for detecting more than five object categories in the corresponding literature. The approaches with None detector indicate the weakly supervised object localization approaches.}
	 \label{table pre-learning} \resizebox{\linewidth}{!}{\begin{tabular}{m{2.5cm}<{\centering}m{1.5cm}<{\centering}m{1.5cm}<{\centering}m{2cm}<{\centering}m{3cm}<{\centering}m{1.5cm}<{\centering}m{3.1cm}<{\centering}m{1.8cm}<{\centering}}
\hline
 \multirow{2}{*}{\textbf{Methods}} &
 \multirow{2}{*}{\textbf{Detector}} &
 \multirow{2}{*}{\textbf{Descriptor}} &
 \multirow{2}{*}{\textbf{Prior knowledge} } &
 \multirow{2}{*}{\textbf{Extra training data}} &
 \multirow{2}{*}{\textbf{Learning model}} &
 \multirow{2}{*}{\textbf{Learning strategy}} &
 \multirow{2}{*}{\textbf{Object category}} \\
 & & & & & & &\\
\hline
\hline
 Li-ISPRS-2018\cite{li2018deep} & CAM* (VGG-F) & CNN & None & ImageNet(tag label) & VGG-F* & Learning learning + CAM learning (patch level) & Remote sensing objects \\
%\hline
 Wilhelem-DICTA-2017\cite{wilhelm2017weakly} & None & CNN & None & ImageNEt(tage label) & CAM & CAM+KDE refine & General objects \\
%\hline
 Tang-TMM-2017\cite{tang2017weakly} & DPM & CNN & Saliency + objectness score & ImageNet(tag label) & DPM+ CNN & Region initialization+DPM and feature learning+bounding box modification & General objects \\
%\hline
 Kolesnikov -BMVC-2016\cite{kolesnikov2016improving} & None & CNN & Human feedback annotation & ImageNet(tag label) & CAM & Active learning for identifying object cluster & General objects \\
%\hline
 Bency-ECCV-2016\cite{bency2016weakly} & None & CNN & None & ImageNet(tag label) & VGG16 & Beam-search based on CNN classifier & General objects \\
%\hline
 %\hline
Zhou-MSSP-2016\cite{zhou2016weakly} & SVM & FC7 & Saliency prior & ImageNet(tag label), remote sensing data(unlabelled) & CNN (AlexNet), SVM & Deep feature transfer +MIL & Remote sensing objects (airplane, car, airport) \\
%\hline
Bergamo-WACV-2016\cite{bazzani2016self} & SVM & CNN & None & ImageNEt(tage label) & CNN, SVM & Mask out initialization + SVM detector training & General objects \\
Hoffman-CVPR-2015\cite{hoffman2015detector} & SVM & FC7 & Detector prior+ representation prior & ImageNet(tag label), ILSVRC13 validation subset(box annotation) & CNN, Latent SVM & Transferring detectors and representation from auxiliary data & General objects \\

\hline
 \end{tabular}}
\end{table*}

\subsection{Initialization and Refinement}
%The approaches in this subcategory focus on developing novel and effective methods for both mining reliable initial instance samples and improving the refinement process for updating the object detectors and the corresponding object locations.
A number of iterative approaches have been developed that take both initialization and refinement into account.
In \cite{siva2011weakly}, Siva et al. propose an intra-class metric and an inter-class metric to initialize the potential object locations.
After obtaining the initial object locations, this method iteratively trains a DPM object detector and uses a model drift detection approach to identify the termination refinement dynamically.
Deselaers et al. \cite{deselaers2010localizing,deselaers2012weakly} present a conditional random field (CRF) model, which is used to learn generic prior knowledge of the objects from meta-training data firstly to localize the potential objects of interest in the weakly labelled training images.
This algorithm updates the CRF model to learn the appearance and shape models for the target object category and localizes the objects of interest in the refinement stage. %
The alternation of localization and learning processes progressively transforms the CRF model from class-generic prior knowledge into the specialized knowledge for a certain target class.
For learning weakly supervised attribute localizer, Wang et al. \cite{wang2013weakly} initialize the learning process by building a Hierarchical Space Tiling (HST) scene configuration model \cite{wang2012hierarchical} and the corresponding appearance models are trained based on HST.
A joint inference and learning process is designed to update the scene attributes and the correlations between the scene parts and attributes gradually.

\subsection{Discussion}
Although the classic weakly supervised learning models are studied in early age, the two-stage learning frameworks, i.e., the learning initialization stage and refinement stage, built by these methods have been widely applied in future works. In the learning initialization stage, these methods provide two kinds of information cues to infer the candidate object regions. The first one is the bottom-up cues, including the region saliency, objectness, intra-class consistency, inter-class discriminability, et al. The other one is the top-down cues, which usually provide the appearance prior for the learning process. Notice that as such top-down cues are hard to obtain from the weakly labeled data, auxiliary training data (with instance-level manual annotation) are usually leveraged to explore the top-down cues which are then transferred to the weakly labeled target data. In the refinement stage, classic machine learning models, such as SVM and CRF, are adopted to gradually refine both the appearance model and locations of the objects of interest.

The advantage of the classic weakly supervised learning methods is that the learning processes can be implemented on small-scale training data and the whole frameworks are quick both in the training phase and the testing phase. The disadvantage is that their performance is not satisfactory, which is due to the limitation in feature representation and model complexity.

\section{Off-the-shelf Deep Models}
\begin{figure}[t]
 \centering
 \includegraphics[width=8.5 cm]{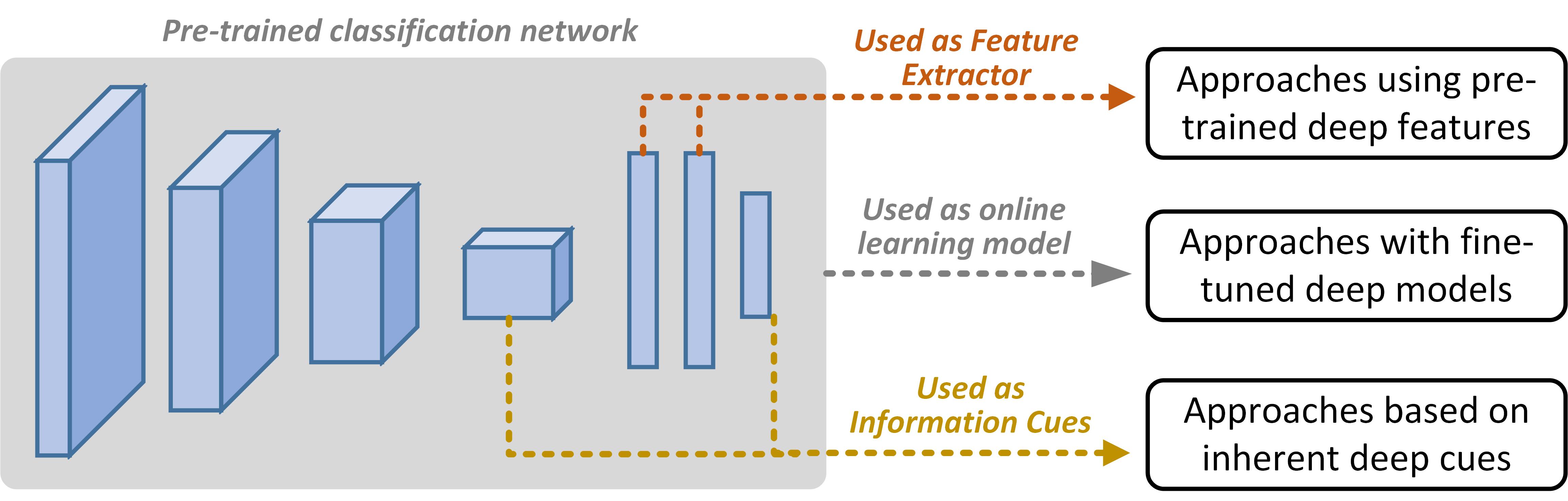}
 \caption{Illustration of different usages of the off-the-shelf deep neural networks by the weakly supervised object localization and detection approaches based on the off-the-shelf deep models. }
 \label{off-the-shelf-flowchart}
 %\vspace{-0.6cm}
\end{figure}

In this section, we review the approaches that learn weakly supervised object localizer or detector based on classic formulations and feature representations based on the deep neural networks, either pre-trained from the ImageNet dataset \cite{russakovsky2015imagenet} (with image tag annotation) or further fine-tuned on the weakly supervised training images in the target domain.
The feature representations are based on the widely used deep models for image classification, such as AlexNet \cite{krizhevsky2012imagenet} and VGG \cite{simonyan2014very}.
The detectors are constructed based on classic formulations such as DPM and SVM~\cite{zhou2016weakly,hoffman2015detector,bazzani2016self,tang2017weakly,ren2016weakly,han2015object,song2014weakly}, or recent models such as RCNN \cite{girshick2014rich} and fast RCNN \cite{girshick2015fast} \cite{shi2017weakly, jie2017deep,uijlings2018revisiting,chen2015webly,kumar2016track}.
%
%MH: need to check this
We further divide these approaches into three subcategories using pre-trained deep features, inherent cues in deep models, and fine-tuned deep models as shown in Fig. \ref{off-the-shelf-flowchart}.

\renewcommand{\multirowsetup}{\centering}
\begin{table*} [t]
 \tiny
	\caption{Summary of the approaches with fine-tuned deep models, which is a subcategory in the weakly supervised object localization and detection approaches based on the off-the-shelf deep models. * indicates a certain variation of the corresponding model. An approach is considered for general object category when it is tested for detecting more than five object categories in the corresponding literature. }
\label{table prepost-learning}.
 \resizebox{\linewidth}{!}{\begin{tabular}{m{2cm}<{\centering}m{2cm}<{\centering}m{1.6cm}<{\centering}m{2.8cm}<{\centering}m{2.5cm}<{\centering}m{1.5cm}<{\centering}m{3.7cm}<{\centering}m{1.5cm}<{\centering}}
\hline
 \multirow{2}{*}{\textbf{Methods}} &
 \multirow{2}{*}{\textbf{Detector}} &
 \multirow{2}{*}{\textbf{Descriptor}} &
 \multirow{2}{*}{\textbf{Prior knowledge} } &
 \multirow{2}{*}{\textbf{Extra training data}} &
 \multirow{2}{*}{\textbf{Learning model}} &
 \multirow{2}{*}{\textbf{Learning strategy}} &
 \multirow{2}{*}{\textbf{Object category}} \\
 & & & & & & &\\
\hline
\hline
 Zhang-IJCV-2019\cite{Zhang2018Leveraging} & Fast RCNN (VGG16) & Pre-trained FC7 & Tag number + mask out prior(AlexNet) & ImageNet(tage label) & SVM & Easy-to-hard & General objects \\
 Uijlings-CVPR-2018\cite{uijlings2018revisiting} & None & CNN & Semantic objectness(SSD*) & ImageNet(tage label), ILSVRC(full annotation) & SSD*+SVM+ Fast RCNN & MIL+knowledge transfer & General objects \\
%\hline
%\hline
 Jie-CVPR-2017\cite{jie2017deep} & Fast RCNN (VGG16) & CNN & Image-to-object transfer prior & ImageNet(tage label) & Fast RCNN (VGG16) & Initialization based on classification network and subgraph discovery + iterative Fast RCNN learning & General objects \\
%\hline
% Li-Arxiv-2017\cite{li2017progressive} & Fast RCNN* (VGG16) & CNN & CAM+mask out prior (AlexNet) +shape prior (grabcut) & ImageNet(tage label) & Fast RCNN* (VGG16), SVM & Progressive representation adaptation & General objects \\
%\hline
Shi-ICCV-2017\cite{shi2017weakly} & Fast RCNN & CNN & Things and stuff prior & ImageNet(tage label), PASCAL Context (full annotation) & FCN,Fast RCNN & Localizing objects based on things and stuff prior and training Fast TCNN iteratively & General objects \\
%\hline
Singh-CVPR-2016\cite{kumar2016track} & Fast RCNN & CNN & Tracking prior & ImageNEt(tage label), Youtube-objects (unlabelled) & Fast RCNN & Discriminative region minning+transferring tracking object pattern + learn object detector & General objects \\
%\hline
Li-CVPR-2016\cite{li2016weakly} & VGG* & CNN & Mask out prior(AlexNet) & ImageNEt(tage label) & VGG*, SVM & Progressive Domain Adaptation & General objects \\
%\hline
Liang-ICCV-2015\cite{liang2015towards} & CNN & CNN & Instance example, motion prior & ImageNEt(tage label) & CNN,R-CNN & Seed selection based on instance example and instance tracking & General objects \\
%\hline
Chen-ICCV-2015\cite{chen2015webly} & RCNN & CNN & Online data type & Web data (weak label) & BLVC net +E-LDA + RCNN & Simple image initialization + graph-based representation adaptation on hard image & General objects \\
%\hline
Zhou-CVPR-2015\cite{zhou2015conceptlearner} & RCNN & FC7 & None & ImageNet(tag label) & SVM, R-CNN & .Max-margin visual concept discovery + Domain-specific detector selection & General objects\\
\hline
 \end{tabular}}
\end{table*}

\subsection{Pre-trained Deep Features}
%One straightforward way to improve the performance of the weakly supervised object localization and detection methods is to exploit the pre-trained deep features.
%
The methods of this category replace the hand-crated feature representations with the pre-trained deep features typically from as AlexNet and VGG.
%
%The real focus of these approaches is the development of novel initialization or refinement methods for WSL.
A brief summary of these approaches is shown in Table \ref{table off-the-shelf feature}.

Song et al. \cite{song2014learning,song2014weakly} determine discriminative feature configurations of an object class via graph modeling, and train object detectors within the multiple-instance learning paradigm.
The deep features of this work are extracted based on the DeCAF scheme~\cite{DonahueDeCAF} and AlexNet~\cite{krizhevsky2012imagenet}.
%
%
%MH: why do you cite these papers? they are not important. %
%Wang et al. \cite{wang2014weakly,wang2015large} use the latent semantic analysis for
%
By using the deep features and spatial features to represent each proposal region, Bilen et al. \cite{bilen2015weakly} propose a convex clustering process for learning the object models under the weak supervision.
The learning objective is able to enforce the similarity among the selected proposal windows.
In \cite{shi2016weakly}, Shi and Ferrari develop a curriculum learning strategy to feed training images into the MIL loop in a pre-defined order, where images containing larger objects are learned at the early stages while images containing smaller objects are learned at later stages.
Ren et al. \cite{ren2016weakly} present a bag-splitting-based MIL mechanism that iteratively generated new negative bags from the positive ones.
This algorithm can gradually reduce the ambiguity in positive images and thus facilitate the learning of more reliable training instance samples.
In \cite{wei2017deep,wei2019unsupervised}, Wei et al. leverage the pre-trained CNN model to implement a Deep Descriptor Transforming process, which can obtain the category-consistent image regions via evaluating the correlations of the descriptors in the convolutional activations of the CNN model.

%MH: what are "inherent deep cues"?
%DW: inherent deep cues indicate the cues existed in the deep network layers inherently rather than those coming from outside of the network. Maybe using this term is not so precise.
%\subsection{Inherent Deep Cues}
\renewcommand{\multirowsetup}{\centering}
\begin{table*} [t]
  \tiny
 	\caption{A brief summary of the approaches using single-network training scheme, which is a subcategory in the weakly supervised object localization and detection approaches with deep weakly supervised learning algorithms. * indicates a certain variation of the corresponding model. An approach is considered for general object category when it is tested for detecting more than five object categories in the corresponding literature. The approaches with None detector indicate the weakly supervised object localization approaches.}
  \label{table endtoend}
  \resizebox{\linewidth}{!}{\begin{tabular}{m{2.5cm}<{\centering}m{1.5cm}<{\centering}m{1cm}<{\centering}m{1cm}<{\centering}m{2.5cm}<{\centering}m{1.5cm}<{\centering}m{2.8cm}<{\centering}m{1.8cm}<{\centering}}
\hline
 \multirow{2}{*}{\textbf{Methods}} &
 \multirow{2}{*}{\textbf{Detector}} &
 \multirow{2}{*}{\textbf{Descriptor}} &
 \multirow{2}{*}{\textbf{Prior knowledge} } &
 \multirow{2}{*}{\textbf{Extra training data}} &
 \multirow{2}{*}{\textbf{Learning model}} &
 \multirow{2}{*}{\textbf{Learning strategy}} &
 \multirow{2}{*}{\textbf{Object category}} \\
 & & & & & & &\\
\hline
\hline
Huang-NIPS-2020\cite{huang2020comprehensive}&Faster RCNN &CNN &None  &ImageNet(tag label) &Faster RCNN*(VGG16/ResNet50) &Proposal attention aggregation and distillation & General objects \\
Shen-CVPR-2020\cite{Shen_2020_CVPR}&Faster RCNN* &CNN &None  &ImageNet(tag label) + Flickr &Faster RCNN*(VGG16) &bagging-mixup + background noise decomposition + clean dat modelling& General objects \\
Mai-CVPR-2020\cite{Mai_2020_CVPR}&None &CNN &None  &ImageNet(tag label) &VGG/InceptionV3 &Integrating discriminative region mining and adversarial erasing & General objects \\
Zhang-ECCV-2020\cite{zhanginter} &None &CNN &Cross-image consistency  &ImageNet(tag label) &VGG/InceptionV3/ ResNet50 &Inter-image stochastic consistency and global consistency & General objects \\
Yang-WACV-2020\cite{Yang_2020_WACV}&None &CNN &None  &ImageNet(tag label) &VGG &Weighted classification activation map combination & General objects \\
Yang-ICCV-2019\cite{yang2019towards} & Faster RCNN* (VGG16) & CNN & None & ImageNet(tag label) & Faster RCNN* (VGG16) + CAM & Online classifier learning with bounding box regression & General objects \\
 Wan-CVPR-2019\cite{wan2019c} & Faster RCNN* (VGG16) & CNN & None & ImageNet(tag label) & Faster RCNN* (VGG16) & Continuation MIL & General objects \\
Shen-CVPR-2019\cite{Shen_2019_CVPR} & WSDDN* (VGG16) & CNN & None & ImageNet(tag label) & Two-stream CNN (WSDDN+DeepLab) & Joint detection and segmentation with cyclic guidance & General objects \\
Wan-CVPR-2019\cite{Wan_2019_CVPR} & Fast RCNN & CNN & None & ImageNet(tag label) & Two-stream CNN & continuation instance selection and detector estimation & General objects \\
Choe-CVPR-2019\cite{Choe_2019_CVPR} & None & CNN & None & ImageNet(tag label) & CNN & Feature learning by attention-based dropout & General objects \\
Jiang-ICCV-2019\cite{jiang2019integral} & None &CNN & None & ImageNet(tag label) & VGG16/Resnet101 & Online attention accumulation on CAM & General objects\\
%Gao-Arxiv-2019\cite{gao2019utilizing} &Faster RCNN* (VGG16) & CNN & None & ImageNet(tag label) & Faster RCNN* (VGG16) & Fusing multiple detection branches in MELM & General objects\\
%Zhou-Arxiv-2019\cite{zhou2019dual} & None & CNN & None & ImageNet(tag label) & CNN & Channel-wise and position-wise attention-based dropout & General objects \\
Sangineto-TPAMI-2018\cite{sangineto2019self} & Fast RCNN (VGG16) & CNN & None & ImageNet(tag label) & Fast RCNN (VGG16) & Easy-to-hard & General objects \\
%\hline
%Jiang-MTA-2018\cite{jiang2018weakly} & WSDDN* (AlexNet) & CNN & None & ImageNet(tag label) & WSDDN* (AlexNet) & WSDDN with attention pooling; multi-scale & General objects \\
%\hline
%Li-Arxiv-2018\cite{li2018weakly} & FCN*(ResNet) & CNN & Saliency prior & ImageNet(tag label) & FCN*(ResNet) & Multi-task FCN for saliency map prediction and image classification & Salient objects \\
%\hline
%Vardazaryan-Arxiv-2018\cite{vardazaryan2018weakly} & None & CNN & None & ImageNet(tag label) & CAM(ResNet) & Class activation mapping & Tools in Laparoscopic videos \\
%\hline
Inoue-CVPR-2018\cite{inoue2018cross} & SSD & CNN & None & PASCAL (full label as source domain),ImageNet & SSD & Domain transfer + pseudo-labeling & Cartoon objects \\
%\hline
 Wan-CVPR-2018\cite{wan2018min} & Faster RCNN* (VGG16) & CNN & None & ImageNet(tag label) & Faster RCNN* (VGG16) & Min-entropy latent modeling & General objects \\
%\hline
Shen-TNNLS-2018\cite{shen2018weakly} & VGG16* & CNN & None & ImageNet (tag label) & vgg16* & Object-specific pixel gradient mapping+Iterative component mining & General objects \\
%\hline
Tang-TPAMI-2018\cite{tang2018pcl} & Fast RCNN* (model ensemble) & CNN & None & ImageNet(tag label) & Fast RCNN* (model ensemble) & MIL+oicr+multi-scale+proposal cluster learning & General objects \\
%\hline
 Zhang-CVPR-2018\cite{zhang2018adversarial} & None & CNN & None & ImageNet(tag label) & VGG16* & Adversarial complementary erasing & General objects (for ILSVRC) \\
%\hline
%Sedai-MLMI-2018\cite{sedai2018deep} & Densenet* & CNN & None & ImageNet(tag label) & Densenet* & Multiscale feature combination, CAM+Attention & Chest Pathologies in X-ray Images \\
%\hline
 Choe-BMVC-2018\cite{choe2018improved} & ResNet & CNN & None & Tiny ImageNet(tag label) & ResNet & GoogLeNet Resize (GR) augmentation & General objects (for Tiny ImageNet) \\
%\hline
 Zhang-ECCV-2018\cite{zhang2018self} & Inception-v3+CAM & CNN & None & ImageNet(tag label) & Inception-v3+CAM & Self-produced guidance learning & General objects (for ILSVRC and CUB) \\
 Gao-ECCV-2018\cite{gao2018c} & Fast RCNN* & CNN & Count (human label) & ImageNet(tag label) & Fast RCNN*+Fast RCNN & WSL with count-based region selection & General objects \\
%\hline
%Grzeszick-Arixv-2018\cite{grzeszick2018weakly} & FCN* (VGG16) & CNN & Point-wise annotation & ImageNet(tag label) & FCN*(VGG16) & Co-occurrence encoding & General objects \\
%\hline
 Singh-ICCV-2017\cite{singh2017hide} & CAM* (GoogLeNet)& CNN & None & ImageNet (tag label) & CAM* (GoogLeNet) & Random hidding patches & General objects (for ILSVRC) \\
%\hline
%Lai-Arxiv-2017\cite{lai2017saliency} & Faster RCNN* (ensamble) & CNN & Saliency prior & ImageNet(tag label) & Faster RCNN* (ensamble) & Seed saliency loss+Seed classification loss+Image loss & General objects \\
%\hline
 Zhu-ICCV-2017\cite{zhu2017soft} & None & CNN & None & ImageNet(tag label) & GoogLeNet* & Soft proposal layer+CAM & General objects \\
%\hline
Wan-ICIP-2017\cite{wan2017weakly} & None & CNN & None & ImageNet(tag label) & CAM*(VGG) & CAM with spatial pyramid pooling layer & General objects \\
%\hline
 Durand-CVPR-2017\cite{durand2017wildcat} & None & CNN & None & ImageNet(tag label) & CAM* (ResNet101) & CAM with multi-map transfer layer & General objects \\
%\hline
Tang-CVPR-2017\cite{tang2017multiple} & Fast RCNN* (model ensemble) & CNN & None & ImageNet(tag label) & Fast RCNN*+Fast RCNN & MIL+oicr+multi-scale & General objects \\
%\hline
 Jiang-CVPR-2017\cite{jiang2017optimizing} & Fast RCNN* (AlexNEt) & CNN & None & PASCAL (edge box),ImageNet & AlexNet+ ROIpool & Region calssification+region selection+multi-scale & General objects \\
%\hline
Diba-CVPR-2017\cite{diba2017weakly} & Faster RCNN* (VGG16) & CNN & None & ImageNet(tag label) & Multi-stream CNN & Cascading LocNet (CAM), SegNet, and MILNet+multi-scale & General objects \\
Selvaraju-ICCV-2017\cite{selvaraju2017grad} & None & CNN & None & ImageNet(tag label) & VGG* & Gradient-based class activation mapping & General objects \\
%\hline
Tang-PR-2017\cite{tang2017deep} & None & CNN & None & ImageNet(tag label) & Fast RCNN* (VGG-16) & SPP with discovery block and classification block & General objects \\
Gudi-BMVC-2017\cite{gudi2017object} & None & CNN & None & ImageNet(tag label) & CAM* (VGG-16) & CAM with Spatial Pyramid Averaged Max (SPAM) Pooling & General objects\\
%\hline
%Rosenfeld-ACCV-2016\cite{rosenfeld2016visual} & None & CNN & None & ImageNet(tag label) & CAM*(VGG) & Recurrent CAM from large to small region & Concepts \\
%\hline
 Bilen-CVPR-2016\cite{bilen2016weakly} & Fast RCNN* (model ensemble) & CNN & None & PASCAL (edge box),ImageNet & Fast RCNN* & MIL+multi scale & General objects \\
%\hline
 Kantorov-ECCV-2016\cite{kantorov2016contextlocnet} & Fast RCNN* (VGG-F) & CNN & Context & ImageNet(tag label) & Fast RCNN* & MIL+multi-scale & General objects \\
%\hline
 Teh-BMVC-2016\cite{teh2016attention} & CNN & CNN & None & PASCAL (edge box),ImageNet & CNN & Proposal attention learning & General objects\\
%\hline
 %Hwang-Arxiv-2016\cite{hwang2016self} & None & CNN & None & None & Two stream CNN & Weighted training classification stream and localization stream with parameter transferring & Tuberculosis \\
%\hline
 Durand-CVPR-2016\cite{durand2016weldon} & None & CNN & None & ImageNet(tag label) & CNN & Feature extraction network+weakly supervised prediction module & General objects \\
%\hline
 Zhou-CVPR-2016\cite{zhou2016learning} & None & CNN & None & ImageNet(tag label) & GoogLeNet* & Class activation mapping & General objects\\
%\hline
 Oquab-CVPR-2015\cite{oquab2015object} & None & CNN & None & ImageNet(tag label) & CNN & Fully convolutional CNN with global max pooling & General objects \\
 %\hline
 Wu-CVPR-2015 \cite{wu2015deep}& None & CNN & None & PASCAL (BING),ImageNet & CNN & Deep multiple instance learning network & General objects\\
\hline
 \end{tabular}}
\end{table*}

\subsection{Inherent Cues in Deep Models}
Instead of using the pre-trained deep models as feature extractor, the methods of this category obtain useful information cues (such as the activations in the intermediate network layers and the semantic scores in the output network layer) from the pre-trained deep neural networks to facilitate the weakly supervised learning process.
The focus of these approaches mainly lies in the initialization stage of the weakly supervised learning process.
A brief summary of these approaches is shown in Table \ref{table pre-learning}.

Bergamo et al. \cite{bazzani2016self} propose a self-taught deep learning approach for localizing objects of interest under weak supervision.
In the initialization stage, they design a mask-out strategy based on the deep semantic cues from a pre-trained classification network.
Specifically, this method first calculates the degeneration of the image-level classification scores when masking out a certain object proposal region and then selects those with large differences as the interested object regions.
After the initialization stage, this method trains an SVM-based object detector in the subsequent refinement stage for final object localization.
Similar to \cite{bazzani2016self}, Bency et al. \cite{bency2016weakly} propose a beam search algorithm to leverage the activation maps of a pre-trained classification network to localize the objects of interest.
This method is based on the observation that when image regions centered around objects of interest are classified by a pre-trained DNN, they obtain higher semantic scores than other image regions.
Hoffman et al. \cite{hoffman2015detector} develop a transfer learning-based algorithm, where the deep neural network is first trained on both the weakly labeled auxiliary training data and the strongly labeled training data to obtain the background detector.
Then, an SVM-MIL model is adopted to learn the object detectors based on the potential foreground regions that are obtained by using the pre-trained DNN to screen the image background regions.
To overcome the problem that the objects of interest would sometimes co-occur with the distracting image background, Kolesnikov et al. \cite{kolesnikov2016improving} {propose a user-guided weakly supervised learning framework to improve the localization capacity.}
This approach first trains a classification network under the image level annotation.
%
%MH: check this sentence. It is not really "active" learning, but rather user in the loop (as the algorithm does not "actively" learn features).
%DW: Agreed
For each image, the intermediate feature maps of the pre-trained network are clustered, and the object clutter is identified by a user.
%
%user is asked to distinguish the object cluster from the background ones.

\subsection{Fine-tuned Deep Models}
The methods of this category fine-tune the off-the-shelf DNN models during the weakly supervised learning process to obtain strong object detectors \cite{xu2020adaptively,chen2015webly,shi2017weakly}.
%
%In other words, these methods not only use the features or other information cues of the pre-trained DNN models but also fine-tune them to obtain the final object detectors.
A brief summary of these approaches is shown in Table \ref{table prepost-learning}.

Chen et al. \cite{chen2015webly} first train CNNs from the web image data via an easy-to-hard learning scheme.
The learned deep features are used to mine object locations by using the exemplar-LDA detector \cite{hariharan2012discriminative}.
The off-the-shelf RCNN detector is then adopted to learn object models based on their localization results.
In \cite{li2016weakly}, Li et al. first use the mask-out strategy based on the pre-trained classification network to obtain the class-specific object proposals.
An SVM-based MIL process is used to localize object instances and
the classification network is further fine-tuned on the localized object instances for better performance.
Shi et al. \cite{shi2017weakly} propose to transfer the prior knowledge of \textit{things} and \textit{stuff} to help the weakly supervised learning process.
A semantic segmentation network is trained from the source set (with available bounding-box annotations) to generate the stuff map and thing map for the weakly labeled images in the target set.
These maps are used to obtain potential object locations, and the fast RCNN model is adopted in a Deep MIL scheme to train the object detectors.
Recently, \cite{uijlings2018revisiting} revisits knowledge transfer for training the weakly supervised object detector.
In this method, a DNN-based proposal extractor is learned from the source data firstly.
The DNN is designed based on the SSD \cite{liu2016ssd} architecture and trained with a semantic hierarchy.
The network is then used to provide proposals and other prior knowledge for the weakly labeled images in the target set.
%
%MH: check this sentence
%DW: CHECKED
An MIL process is used to determine the proposals that cover the objects of interest
based on which the fast RCNN model is adopted to learn the final object detectors.
In \cite{Zhang2018Leveraging}, Zhang et al. first learn to localize the objects of interest via a collaborative self-paced curriculum learning mechanism based on pre-trained deep features.
The fast RCNN model is applied to learn object detectors.
%based on the localization results.

\renewcommand{\multirowsetup}{\centering}
\begin{table*} [t]
  \tiny
	\caption{A brief summary of the approaches with multi-network training, which is a subcategory in the weakly supervised object localization and detection approaches with deep weakly supervised learning algorithms. * indicates a certain variation of the corresponding model. An approach is considered for general object category when it is tested for detecting more than five object categories in the corresponding literature. The approaches with None detector indicate the weakly supervised object localization approaches.}
	\label{table multinet} \resizebox{\linewidth}{!}{\begin{tabular}{m{2cm}<{\centering}m{1.5cm}<{\centering}m{1.5cm}<{\centering}m{1.8cm}<{\centering}m{3cm}<{\centering}m{2.6cm}<{\centering}m{3cm}<{\centering}m{1.5cm}<{\centering}}
\hline
 \multirow{2}{*}{\textbf{Methods}} &
 \multirow{2}{*}{\textbf{Detector}} &
 \multirow{2}{*}{\textbf{Descriptor}} &
 \multirow{2}{*}{\textbf{Prior knowledge} } &
 \multirow{2}{*}{\textbf{Extra training data}} &
 \multirow{2}{*}{\textbf{Learning model}} &
 \multirow{2}{*}{\textbf{Learning strategy}} &
 \multirow{2}{*}{\textbf{Object category}} \\
  & & & & & & &\\
\hline
\hline
Zhang-CVPR-2020\cite{Zhang_2020_CVPR} &None &CNN &Common object co-localization &ImageNet(tag label) &VGG/InceptionV3/ResNet50/ DenseNet161 &Classification + pseudo supervised object localization & General objects \\
Zhong-ECCV-2020\cite{zhong2020boosting} &Faster RCNN &CNN &Location prior &ImageNet(tag label) + COCO (box label) &Oneclass universal detector + MIL classifier (on ResNet50) &Progressive knowledge transfer & General objects \\
Kosugi-ICCV-2019\cite{kosugi2019object} & Fast RCNN* &CNN & Mask-out prior & ImageNet(tag label) & Mask-out net + OICR* & Mask-our prior-guided label refinement & General objects \\
Singh-CVPR-2019\cite{ Singh_2019_CVPR} & Fast RCNN* & CNN & Motion prior & ImageNet(tag label), videos & RPN+WSDDN/OICR (VGG16) & Training RPN using weakly-labeled videos for WSOD & General objects \\
Arun-CVPR-2019\cite{ Arun_2019_CVPR} & Fast RCNN & CNN & None & ImageNet(tag label) & Fast RCNN (VGG16) + Fast RCNN* (VGG16) & Employ dissimilarity coefficient for modeling uncertainty & General objects \\
Li-TPAMI-2019\cite{li2018mixed} & Faster RCNN* (VGG16) & CNN & Objectness (classifier) prior & ImageNet(tag label), ILSVRC2013(box label for unseen categories) & Faster RCNN* (VGG16) *2 & Objectness transfer+MIL +multi-scale & General objects \\
%\hline
%Li-Arxiv-2019\cite{li2019weakly} & Fast RCNN* & CNN & None & ImageNet(tag label) & OICR (VGG16) +CPN\cite{chen2018cascaded} + ResNet101 & Online collaboration of the segmentation network and detection network to generate the pseudo ground-truths for both &General objects \\
 Zhang-ECCV-2018\cite{zhang2018ml} & fast RCNN* (VGG16) & CNN & None & PASCAL (edge box), ImageNet(tag label) & Multi-view WSDDN+multi view Fast RCNN & Two phase multi-view learning & General objects \\
%\hline
Shen-CVPR-2018\cite{shen2018generative} & SSD & CNN & None & ImageNet(tag label) & SSD+Fast RCNN* & MIL+GAN & General objects \\
%\hline
 Zhang-CVPR-2018\cite{zhang2018w2f} & Faster RCNN (VGG16) & CNN & None & ImageNet(tag label) & MIDN+Faster RCNN & WSOD+Pseudo Ground-truth Mining+FOD & General objects \\
%\hline
Zhang-CVPR-2018\cite{zhang2018zigzag} & Fast RCNN* (VGG16) & CNN & None & PASCAL (edge box), ImageNet & WSDDN + Fast RCNN* & WSDDN+easy-to-hard FOD & General objects \\
%\hline
Tang-ECCV-2018\cite{tang2018weakly} & Fast RCNN* (VGG16) & CNN & None & ImageNet(tag label) & Fast RCNN* (VGG16) & Alternating training of WSRPN and WSOD+multi-scale & General objects \\
%\hline
Tao-TMM-2018\cite{tao2018exploiting} & Fast RCNN* (VGG16) & CNN & Web image & Web dataset(weak label),imageNet(tag label) & Midn & Easy-to-hard & General objects \\
%\hline
 %Arun-Arxiv-2018\cite{arun2018dissimilarity} & Faster RCNN* (model ensemble) & CNN & Distribution consistency prior & ImageNet(tag label) & Faster RCNN* (model ensemble) & Dissimilarity coeffcient minimialization & General objects \\
%\hline
 %Bartz-Arxiv-2018\cite{bartz2018loans} & ResNet* & CNN & None & ImageNet(tag label) & ResNet* & Student-teacher fashion & Sheep, figure skating \\
%\hline
 Wang-IJCAI-2018\cite{wang2018collaborative} & Faster RCNN (VGG16) & CNN & Model consistency & ImageNet(tag label) & Faster RCNN+Fast RCNN* & MIL+collaborative learning & General objects \\
%\hline
Wei-ECCV-2018\cite{wei2018ts2c} & Faster RCNN* (VGG16) & CNN & Shape prior+ context prior & ImageNet(tag label) & MIDN+CAM+ Deeplab & Tight Box Mining+MIL +OICR+multi-scale & General objects \\
%\hline
 Ge-CVPR-2018\cite{ge2018multi} & Faster RCNN (VGG16) & CNN & Local objectness and global attention & ImageNet(tag label) & MIDN,TripNet, GoogleNet, FCN, Fast RCNN & Multi evidence fusion+ outlier filtering +pixel label preidction +box generation+multi-scale & General objects \\
%\hline
%Li-ICME-2018\cite{li2018sfcm} & None(only annotator) & CNN & None & ImageNet(tag label) & VGG16* & Global learnable pooling+selective feature category mapping & General objects \\
%\hline
%Zheng-PRICAI-2018\cite{zheng2018weakly} & None(only annotator) & CNN & None & ImageNet(tag label) & VGG16*+DRL & Deep reinforcement learning for cutting background & General objects \\
%\hline
%Kim-ICCV-2017\cite{kim2017two} & None(only annotator) & CNN & None & ImageNet(tag label) & CAM* *2 & Suppressing the discriminative local parts & General objects \\
%\hline
% Yan-Arxiv-2017\cite{yan2017weakly} & Faster RCNN (VGG16) & CNN & None & PASCAL (edge box), ImageNet(tag label) & WSDDN + Fast RCNN* & Expectation-maximization learning & General objects \\
%\hline
Dong-MM-2017\cite{dong2017dual} & Fast RCNN* & CNN & None & ImageNet(tag label) & Fast RCNN*+R-FCN & Easy-to-hard & General objects \\
%\hline
 Li-BMVC-2017\cite{li2017multiple} & Fast RCNN* & CNN & Shape prior & ImageNet(tag label) & Fast RCNN* +CAM+DeepLab & Easy-to-hard & General objects \\
%\hline
Wang-CVPR-2017\cite{wang2017learning} & CNN & CNN & None & ImageNet(tag label) & CAM* & Image-level training+pixel-level fine tuning & Salient objects \\
%\hline
 %
%\hline
%Yang-Arxiv-2016\cite{yang2016weakly} & Fast RCNN & CNN & None & ImageNet(tag label) & WSDDN + label denosiying+ Fast RCNN & Cascade learning & General objects \\
%\hline
 Sun-CVPR-2016\cite{sun2016pronet} & None & CNN & None & ImageNet(tag label) & Multi-scale FCN+ CNN(vgg16) & Cascade localization and recognition & General objects \\
%\hline
 Zhang-TGRS-2016\cite{zhang2016weakly} & CNN & CNN & None & ImageNet(tag label), auxiliary data(image label) & CPRNet+LocNet & Alternative training CPRNet and LocNet & Aircraft \\
\hline
 \end{tabular}}
\end{table*}

\subsection{Discussion}
Introducing the off-the-shelf deep models into the weakly supervised object localization or detection framework is the most straightforward approach to integrate deep learning and weakly supervised learning.
%
%Although the ideas are usually simple and straightforward, appreciate performance gain is obtained when compared these methods with the classic models.
%
The methods in this category show that: 1) feature learning is an important factor to improve the weakly supervised learning process; 2) DCNN models can infer discriminative spatial locations when learned under the image-level supervision; 3) pre-training DNN models on large-scale auxiliary training data is a simple but effective way to encode useful cues for the weakly supervised learning process.
Compared with the classic models,
the methods of this category exploit large-scale auxiliary training data to learn powerful feature representations and top-down cues.
By using DNN models as the object detector or localizer, a significant performance gain can be obtained.
However, more effective feature learning models can be exploited in the weakly supervised learning process.

\section{Deep Weakly Supervised Learning}
In this section, we review the methods that learn weakly supervised object localizers or detectors by designing novel deep weakly supervised learning frameworks\footnote{Notice that here we mainly indicate that the core learning blocks used in the learning frameworks are based on deep learning, while some minor computational components in pre-processing or post-processing, such as proposal extraction and bounding-box modification, et al., are not necessarily be implemented by deep learning.}.
Different from the approaches discussed in previous sections, both the feature representations and the object detectors of the approaches in this category are learned by newly-designed deep neural networks.
The whole weakly supervised learning framework may be designed in a compact network model, such as in \cite{zhu2017soft,oquab2015object,kantorov2016contextlocnet,selvaraju2017grad,Gao_2019_cmidn,Xue_2019_ICCV,zhou2019dual,Yang_2020_WACV,mai2020erasing,zhang2020progressive}, or contain several function-distinct DNN components, such as in \cite{zhang2018ml,wei2018ts2c,diba2017weakly,wang2017learning,sun2016pronet,zhang2020rethinking,lu2020geometry}.
We categorize these approaches into two groups using single-network training and multi-network training, respectively.
%
%More concrete review of the approaches in each subcategory will be described below.

\subsection{Single-Network Training}
The methods of this category are designed with a single deep neural network using the training images (or together with the extracted object proposals) as inputs and the image-level classification labels as the outputs.
These approaches do not usually rely on meticulously designed initialization processes to obtain the potential object regions.
Instead, these methods discover the interested object regions solely based on the end-to-end learning process of the designed DNN models.
The DNN models used in these approaches usually have similar feature learning layers as the conventional image classification network, e.g., AlexNet, VGG, GoogleNet, and ResNet, followed by the instance label inferring and image label propagation layers to inherently predict the labels of each proposal region and generate the final image labels from the predicted proposal labels, respectively. Some of the methods contain multiple network streams for online inferring multiple informative cues.
A brief summary of these approaches are
shown in Table \ref{table endtoend}.

%Early efforts in this direction mainly explore ways to implement MIL in DNN model, i.e., the deep MIL.
%
In the DNN models proposed by Wu et al. \cite{wu2015deep} and Bilen et al. \cite{bilen2016weakly}, the network inputs are the training images and extracted object proposal regions while the outputs are the image-level semantic scores.
The first parts of these networks extract the features and infer the labels for each proposal region, and the second parts propagate the proposal scores to the image-level via the max pooling scheme \cite{wu2015deep} or the two-stream score regularization method \cite{bilen2016weakly}.
Zhou et al. \cite{zhou2016learning} present an end-to-end weakly supervised deep learning based on the class activation mapping (CAM).
The weights of the feature maps in the intermediate layers are inferred based on the correspondence between the feature map and a certain object category.
The feature maps are then combined to form the class activation maps based on the inferred weights, which highlight the locations of the objects of interest. {Notably, this method is determined to be highly efficient in recent work \cite{choe2020evaluating}. }
Built on CAM \cite{zhou2016learning}, Durand et al. \cite{durand2017wildcat} introduce the multi-map transfer layer and the WILDCAT pooling layer to facilitate the more accurate deep MIL process.

Recently, a number of two-branch MIL models have been developed in which one is based on a typical deep network and the other one is introduced for weakly supervised learning.
Based on the WSDDN \cite{bilen2016weakly}, Tang et al. \cite{tang2017deep} integrate MIL branch and the instance classifier refinement branch into a unified deep learning framework such that more accurate online instance classifier learning is realized under the weak supervision.
In \cite{diba2017weakly}, Diba et al. propose a weakly supervised cascaded convolutional network, which contains three branches.
The first branch adopts the CAM module to generate the class activation maps. %
The second branch uses the generated class activation maps as the supervision signal to learn a segmentation module to generate the segmentation masks of the objects of interest.
Using the candidate object proposals selected based on the obtained segmentation masks as supervision, the third network branch uses a MIL process to mine accurate object locations from the candidate object regions.
In \cite{zhang2018adversarial}, Zhang et al. present a CAM-based network architecture which contains a classification branch and a counterpart classifier branch for object localization.
Specifically, the classification branch is used to localize the discriminative object regions, which drives the counterpart classifier branch to discover new and complementary object regions by erasing its discovered regions from the feature maps.
Wan et al. \cite{wan2018min} propose a min-entropy latent model (MELM) for weakly supervised object detection based on the assumption that minimizing entropy results in minimum randomness of a system.
The network architecture is similar to \cite{tang2017deep}, but global min-entropy and local min-entropy losses are introduced to train a DNN model to select the proposal cliques of largest object probability and mine truthful object locations from the selected proposal cliques.
Zhang et al. \cite{zhang2018self} develop a self-generated guidance method for weakly supervised object localization.
%
%MH: check this sentence
%DW: CHECKED
In this work, a self-generated guidance map is derived from a CAM {layer to help learning features and object location masks from the previous network layers.}
 {More recently, Gao et al.\cite{gao2021ts} propose a token semantic coupled attention mapping for WSOL, which models the long-range visual dependency of the image regions and thus avoid partial activation.  Ren et al.\cite{ren2020instance} introduce the instance-associative spatial diversification constraints and build the parametric spatial dropout block to address the instance ambiguity and incomplete localization problems. Besides, they additionally adopt a sequential batch back-propagation algorithm, which enables their model to use a large ResNet as the backbone\footnote{{As discussed in \cite{Enabling2020}, it is non-trivial to introduce the deeper network backbones, e.g, the deep residual network, into the weakly supervised object detection frameworks as it would encounter dramatic deterioration of detection accuracy and training non-convergence. }}. Although there are other methods that use ResNet as the backbone \cite{yun2019cutmix,Choe_2019_CVPR,baerethinking}, there is very limited exploration of using more recent backbone architectures, e.g., DesNet \cite{huang2018condensenet} and Res2net \cite{gao2019res2net}, in both WSOL and WSOD frameworks. }

\subsection{Multi-Network Training }
The methods in this school collaborate multiple networks, either in one training stage or in multiple training stages, to accomplish the weakly supervised object localization or detection task.
The approaches of this category usually train a network to mine the initial object regions \cite{li2018mixed,wei2018ts2c,ge2018multi} and another network for the detection task under the MIL framework \cite{li2018mixed,zhang2018zigzag,tang2018weakly,wang2018collaborative}.
An additional object detection network, e.g., Fast RCNN, may also be used to train the final object detectors \cite{zhang2018w2f,ge2018multi,zhang2018learning}.
By integrating multiple networks, these methods tend to achieve better performance both in object localization and detection.
A brief summary of these approaches is shown in Table \ref{table multinet}.

Li et al. \cite{li2017multiple} propose a multiple instance curriculum learning method, where a network based on the WSDDN \cite{bilen2016weakly} model is used to mine candidate object proposals and another one based on the CAM \cite{zhou2016learning} algorithm to generate saliency maps from the selected proposals.
A curriculum is designed to select confident training examples based on the consistency between the regions outputted by the two networks.
The object detectors are then trained by using the confident training examples iteratively.
Dong et al. \cite{dong2017dual} present a dual-network progressive approach for weakly supervised object detection, where a positive instance selection network and a region refinement network are adopted to minimize the classification error and modify object localization, respectively.
These two networks are worked under a co-training paradigm.
In \cite{ge2018multi}, Ge et al. first obtain intermediate object localization and pixel labeling results using a classification network.
A triplet-loss network and an instance classification network are then constructed to detect outlier and filter object instances.
Finally, the filtered object instances are used as the supervision to train another Fast RCNN-based detection network.
 {In order to overcome the limitations brought by the imprecise of the extracted object proposals,} Wei et al. \cite{wei2018ts2c} propose to mine object proposals with tight boxes to learn weakly supervised object detector. {The assumption is that the proposals with tight boxes are more likely to contain the objects of interest thus mining such kind of proposals would help screen the cluttered background regions. In their approach, }
a semantic segmentation network is first learned using the object localization map generated by CAM as the pseudo ground-truth.
The predicted segmentation masks are used to mine object proposals with tight boxes, and fed into the online instance classifier refinement (OICR) network to learn weakly supervised object detector.
%
%MH: how do they train OICR network?
 {With the same motivation as \cite{wei2018ts2c}, Tang et al. \cite{tang2018weakly} propose to combine a two-stage region proposal network with an OICR network to learn the weakly supervised object detectors. Instead of mining proposals based on the semantic segmentation cue, Tang et al. use both the low-level cues (feature maps in shallow layers) and the high-level cues (semantic scores in deep layers) to mine reliable proposals in the two stages, respectively. The parameters of the region proposal network are obtained based on the network trained by \cite{tang2017deep}.}
 {Zhang et al.\cite{zhang2018zigzag} explore the reliability of each training image by evaluating the image difficulty and then feed the images into the learning procedure in an easy-to-hard order. Specifically, the image difficulty is evaluated by diagnosing the localization outputs of the pre-trained WSDDN model based on the concentrateness of the high-scored proposal locations. }
In \cite{zhang2018w2f}, Zhang et al. use three networks to learn weakly supervised object detectors.
%
%MH: how do they learn? or they just OICR network?
%DW: they just use the OICR network directly..
First, an OICR network is trained to generate the initial object regions.
%
%MH: do they train or just use RPN?
%DW: train on pseudo ground-truth data
Then, they train an RPN \cite{Ren2017Faster} based on {the pseudo ground-truth boxes obtained after implementing a post-process on the initial object regions, and use the learned RPN} to generate more accurate object locations.
%MH: what is it?
Finally, fully supervised object detectors \cite{girshick2015fast,Ren2017Faster} are trained based on the obtained object locations.

%MH: your writing is rally mechanical. You often say
%In [X], some people propose.... First, .... Then, .... Finally, ...
%This is really boring and does not bring any sufficient insight. A good survey paper needs to discuss and summarize the most work with insight. NOT just description of what has been done. Your writing is like taking some sentences from abstract and describing each work in three or four sentences. What are the connections with other methods? Why are they important and useful? Those are important things to describe (not just 3/4 sentences without substantial details).

\subsection{Discussion}
Compared with the off-the-shelf deep model-based weakly supervised object detection and localization methods, the deep weakly supervised learning methods exploits the merits of deep learning and weakly supervised learning approaches.
Without complex design in the learning initialization stage, the end-to-end deep weakly supervised learning methods have been shown to perform well by introducing the MIL mechanism into the network design of the DCNN models.
While the multi-network training methods can further improve the learning performance by combining multiple function-specific networks.
%
%The advantage of the methods of this category is that the deep learning process and the weakly supervised learning process are integrated into a unified learning framework, leading to superior capacity in feature learning and object appearance modeling.
%
%One limitation is that these approaches are mostly implemented in the data-driven fashion, whereas the weakly labeled data in the weakly supervised learning process are not as reliable as those in the fully supervised learning process. Thus, helpful prior knowledge cues that may guide the deep weakly supervised learning process are still under-studied.
On the other hand, the performance of these methods is limited by whether the information extracted from the weakly-supervised module is effective or not.
As such, prior knowledge may be useful to guide the deep learning process without solely relying on the weakly-supervised network.

\section{Datasets and Evaluation Metrics}
During the last decades, significant efforts have been made to develop various methods for learning weakly supervised object localizer or detector.
For fair performance evaluation, it is of great importance to introduce some publicly available benchmark datasets and evaluation metrics.
%
%Here we summarize the widely used open-published datasets and the standard evaluation metrics.
\begin{table}[t]
  \centering
  \caption{ {Brief summarization of the characteristics of the datasets. The top three are the datasets usually used for the WSOD task, while the bottom two are the common datasets for the WSOL task. \#Categories indicates the number of object categories. \#Images indicates the number of images. ``GO'' is short for Generic Objects. }}
    \begin{tabular}{ccccc}
    \hline
    Dataset & \multicolumn{1}{c}{Content} & \multicolumn{1}{c}{\#Categories} & \multicolumn{1}{c}{\#Images} & \multicolumn{1}{c}{Metrics} \\
    \hline\hline
    PASCAL VOC 07 & \multirow{3}[2]{*}{GO} & \multirow{3}[2]{*}{20} & 9,962 & \multirow{3}[2]{*}{mAP, CorLoc} \\
    PASCAL VOC 10 &       &       & 21,738 &  \\
    PASCAL VOC 12 &       &       & 22,531 &  \\
    \hline
    CUB-200-2011   & Birds & 200   & 11,788 & Top-1/5 Loc \\
    ILSVRC 2016 & GO    & 1000  & 1.2 M & GT Loc \\
    \hline
    \end{tabular}%
  \label{tab:dataset}%
\end{table}%

\begin{figure}[t]
 \centering
 \includegraphics[width=9.5 cm]{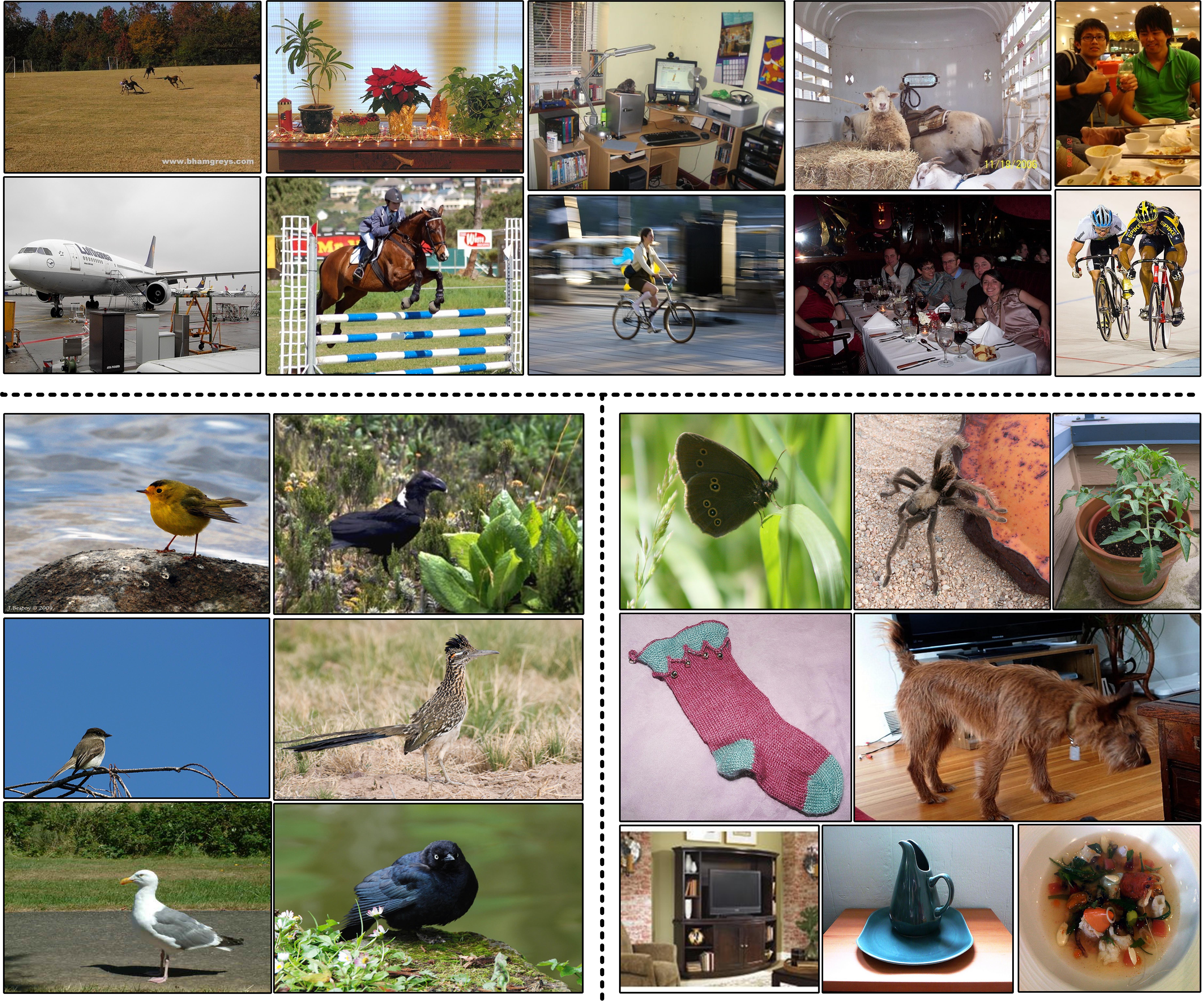}
 \caption{ {Illustration of examples from the PASCAL VOC (top block), CUB-200-2011 (bottom-left block), and ILSVRC 2016 (bottom-right block) datasets.}}
 %\vspace{-0.6cm}
 \label{dataset}
\end{figure}

Existing weakly supervised object detection methods are usually evaluated on the PASCAL VOC datasets, including the PASCAL VOC 2007, 2010, and 2012 sets.
The PASCAL VOC 2007 \cite{everingham2007pascal}, PASCAL VOC 2010 \cite{everingham2010pascal} and PASCAL VOC 2012 \cite{everingham2015pascal} contain 9,962, 21,738, and 22,531 images of 20 object classes.
These three datasets are divided into train, val, and test sets, where the trainval set (5,011 images for PASCAL VOC 2007, 10,869 images for PASCAL VOC 2010, and 11,540 images for PASCAL VOC 2012) are used to train the weakly supervised object detector, and the rest for evaluation.
%while the test set (4,951 images for PASCAL VOC 2007, 10,869 images for PASCAL VOC 2010, and 10,991 images for PASCAL VOC 2012) are used to test the object detection performance.
%
The mean of AP (mAP) metric is used to measure the performance where one object is successfully detected if the intersection over union (IoU) between the ground-truth and predicted boxes is more than 50 percentage.

The weakly supervised object localization performance is usually evaluated on the PASCAL VOC, ILSVRC, and CUB datasets.
On the PASCAL VOC datasets \cite{everingham2007pascal,everingham2010pascal,everingham2015pascal}, weakly supervised object localization methods only use the trainval sets, which are different from the setting in weakly supervised object detection.
That is, both the weakly supervised learning process and localization process are implemented on the same image data.
To evaluate the localization performance on the PASCAL VOC datasets, the correct localization metric (CorLoc) is adopted, where the bounding-box with the highest class-specific score from each image is examined to be whether correct (with more than 50\% overlap with the ground-truth box) or not.
In addition to the PASCAL VOC datasets, the ILSVRC 2016 dataset \cite {russakovsky2015imagenet} (i.e., the ImageNet) and CUB-200-2011 dataset \cite{wah2011caltech} are also widely used for performance evaluation.
The ILSVRC 2016 dataset contains more than 1.2 million images of 1,000 classes for training, while the validation set, which contains 50,000 images, is used for testing.
The CUB-200-2011 dataset contains 11,788 images of 200 categories with 5,994 images for training and 5,794 for testing.
%
%MH: ? Is it trivial to state this?
%On these two datasets, the training set and the test set are different image sets, and the evaluation metric is top-1 or top-5 localization error.
%
For these two datasets, {The commonly-used evaluation metrics are GT-known localization accuracy (GT Loc), Top-1 localization accuracy (Top-1 Loc), and Top-5 localization accuracy (Top-5 Loc). Specifically, GT Loc judges the localization results as correct when the intersection over union (IoU) between the ground-truth bounding box and the estimated box is no less than 50\%, while Top-1 Loc considers the localization results as correct when the class predicted with the highest score is equal to the ground-truth class and the estimated bounding box has no less than 50\% IoU with the ground-truth bounding box \cite{Choe_2019_CVPR}. Top-5 Loc differs from Top-1 Loc in that it checks if the target label is one of the top 5 predictions. As can be seen, the Top-1 Loc is a harder metric than the GT Loc as it needs to additionally predict the class label correctly. This will dramatically increase the task difficulty when performing under very large or fine-grained semantic spaces. The difficulty of Top-5 Loc is between Top-1 Loc and GT Loc as it requires the model to predict the class label but does not restrict the prediction to be perfectly correct.}

 {We provide a brief summarization of the characteristics of the aforementioned datasets in Table \ref{tab:dataset}. We additionally show some examples from each dataset to illustrate the bias of image content in different datasets. As displayed in Fig. \ref{dataset}, the PASCAL datasets contain relatively more complex image content, where multiple object instances and categories may appear in a single image and different images would contain objects with significant scale variations. Although it only contains 20 object categories, its category diversity is higher than the CUB-200-2011 dataset as all the 200 categories in the CUB-200-2011 dataset are related to birds. ILSVRC 2016 contains far more images and categories than the PASCAL datasets. However, the content of the images from the ILSVRC 2016 dataset tends to be simpler than that of the PASCAL datasets---each of the images from the ILSVRC 2016 dataset typically contains only one single object and the objects have more consistent sizes and are placed in clearer background context relative to those from the PASCAL datasets. }\\

%MH: All these applications are not main stream vision tasks. I am sure reveiwers will complain why you discuss these methods that are not of great importance to the vision community (except video and medical). Even if I swap the order, they will still complain.
%DW: This section is meant to summarize the application directions of the current weakly supervised object detection or localization methods. Although some directions are not of great importance to the vision community, they are indeed the application directions of the existing weakly supervised object detection or localization methods. And these applications may also inspire readers from other research communities.

\begin{figure}[t]
 \centering
 \includegraphics[width=9.5 cm]{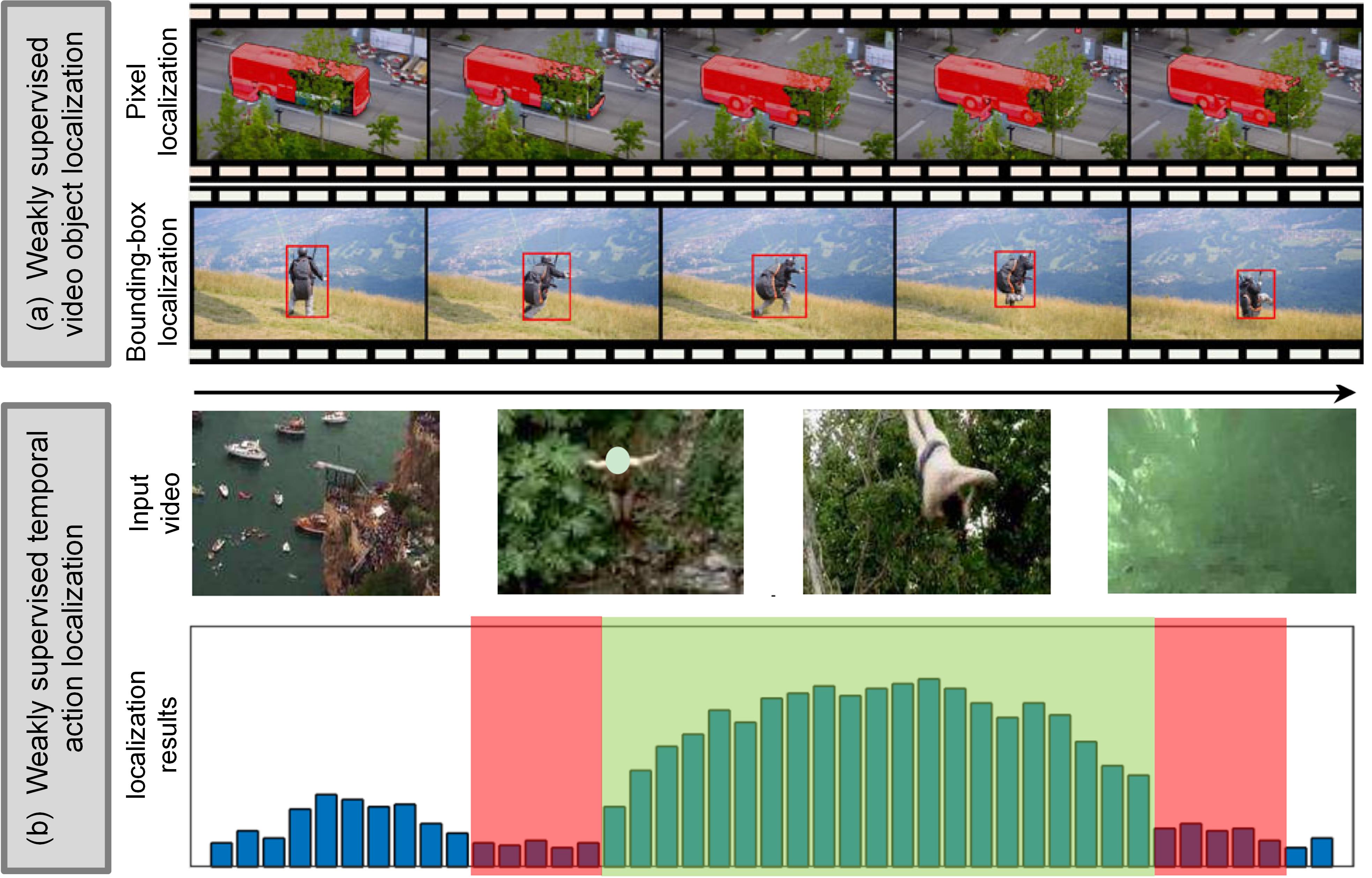}
 \caption{Weakly supervised object localization or detection methods for video understanding. The examples are from \cite{zhang2018spftn,shou2018autoloc}. }
 \label{videoap}
 %\vspace{-0.6cm}
\end{figure}

\section{Applications}
In recent years, weakly supervised object localization and detection techniques have been used in numerous vision problems, especially when it is difficult to collect ground-truth labels.
%
%By summarizing the existing literature, we find that the common property of these applications is that their manually-annotated ground-truth labels are very hard to obtain and these applications can be roughly divided into the following five directions.

\subsection{Video Understanding}
As it is time-consuming to obtain object-level annotations for each video frame, weakly supervised object localization and detection methods have also been applied in the field of video understanding \cite{chanda2017adopt,zhang2018spftn,shou2018autoloc,huh2019patch,yuan2019marginalized,schroeter2019weakly,ma2020sf} (see Fig. \ref{videoap}).
For example, Chanda et al. \cite{chanda2017adopt} build a two-stream learning framework, which adapts the information from the labeled images (source domain) to the weakly labeled videos (target domain).
In \cite{zhang2018spftn} Zhang et al. propose a self-paced fine-tuning network for learning two network heads to localize and segment the object of interest from the weakly labeled training videos.
The network is equipped with the multi-task self-paced learning function which can integrate confident knowledge from each single task (localization or segmentation) and use it to build stronger deep feature representation for both tasks.
On the other hand, \cite{shou2018autoloc,singh2017hide,wang2017untrimmednets,nguyen2018weakly} develop methods to localize temporal actions in the given untrimmed videos, where the main goal is to predict the temporal boundary of each action instance contained in the weakly labeled training videos.
 {Essentially, such a weakly supervised action localization (WSAL) task is an emerging, yet rapidly developing topic in recent years, and the methods for solving this task are highly related to the weakly supervised object detection and localization methods. The additional challenges are: (i) the duration of the interesting action has very large variation, i.e., from a few seconds to thousands of seconds; and (ii) the features extracted to represent the interesting action would be entangled with those of the complex scenes of the video frame. }
 {Notice that when applying to video understanding, there are strong correlations among adjacent video frames. So, additional informative constraints can be introduced to facilitate the weakly supervised object detection or localization under this scenario. }

%MH: when you take images from other people, did you ask them for permission?
%DW: Yes, have asked.

\subsection{Art Image Analysis}
\begin{figure}[t]
 \centering
 \includegraphics[width=9.5 cm]{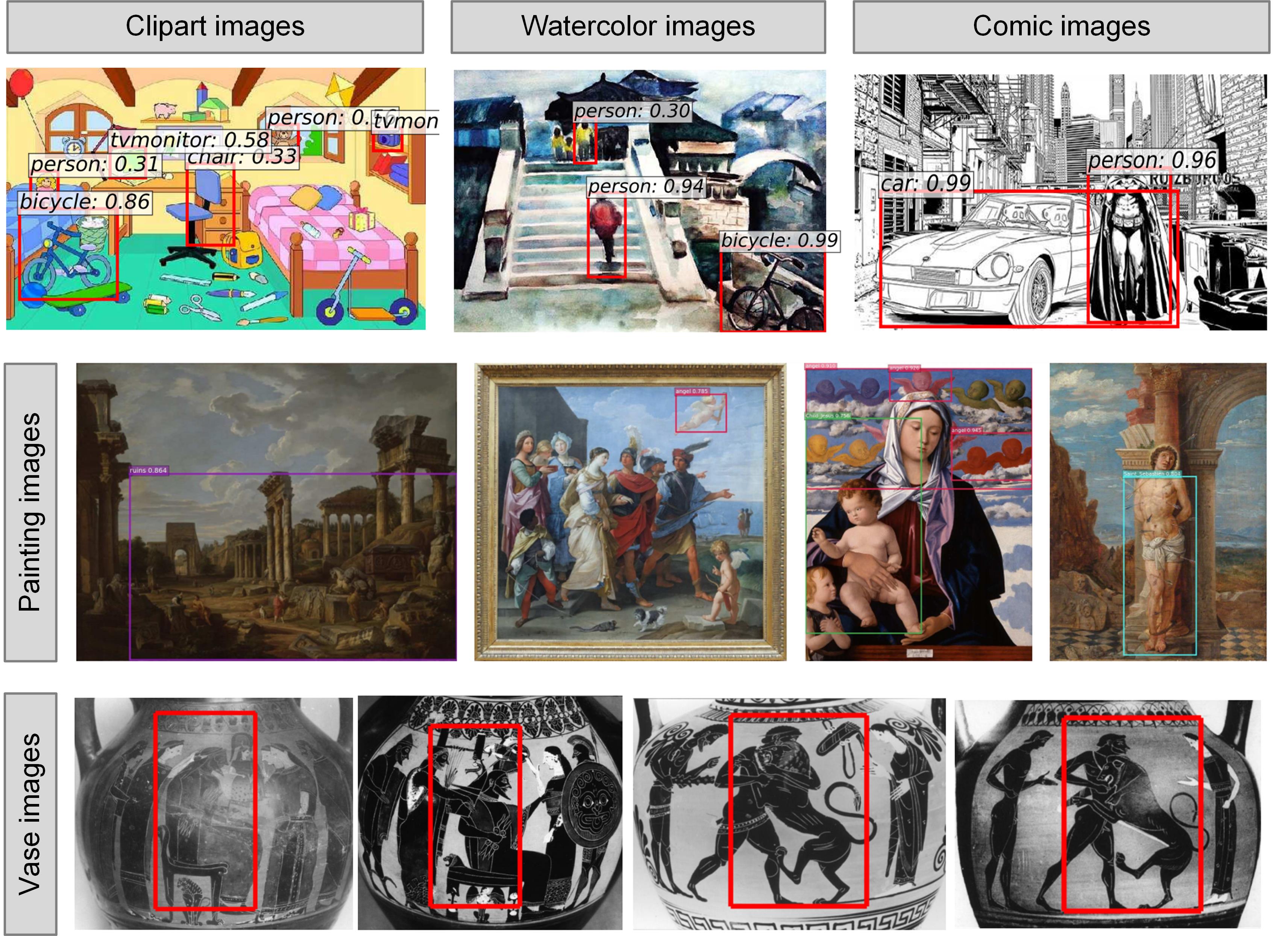}
 \caption{Examples of the application of weakly supervised object localization or detection approaches for the analysis of art images.
 The examples are from \cite{inoue2018cross,gonthier2018weakly,crowley2013gods}, where detection results in different colors in the painting images indicate different types of objects. }
 \label{art}
 %\vspace{-0.6cm}
\end{figure}

One interesting application of the weakly supervised object localization and detection techniques is the analysis of art images (see Fig. \ref{art}).
Inoue et al. \cite{inoue2018cross} propose a cross-domain weakly-supervised object detection framework for learning the object detectors from weakly labeled watercolor images.
A progressive domain adaptation method to transfer the style of the fully-labeled data from the source domain (the normal RGB domain) to the target domain (the watercolor domain) is developed.
In \cite{gonthier2018weakly} Gonthier et al. propose a weakly supervised learning algorithm for detecting objects in paintings.
The IconArt database which contains object classes that are absent from the photographs in daily life is developed for performance evaluation.
In addition, Crowley
and Zisserman \cite{crowley2013gods} adopt a weakly supervised object localization scheme for automatically annotating images of gods and animals in decorations on classical Greek vases.
 {When applying to art image analysis, a key challenge arises due to the distinctiveness of the content domain---even the same semantics and image scenes would be presented differently to those in the natural environment. Under this scenario, models with stronger self-domain adaptation capacity would be required for the task.}

%MH: when you take images from other people, did you ask them for permission?
%DW: Yes, have asked.

\subsection{Medical Imaging}
\begin{figure}[t]
 \centering
 \includegraphics[width=9.5 cm]{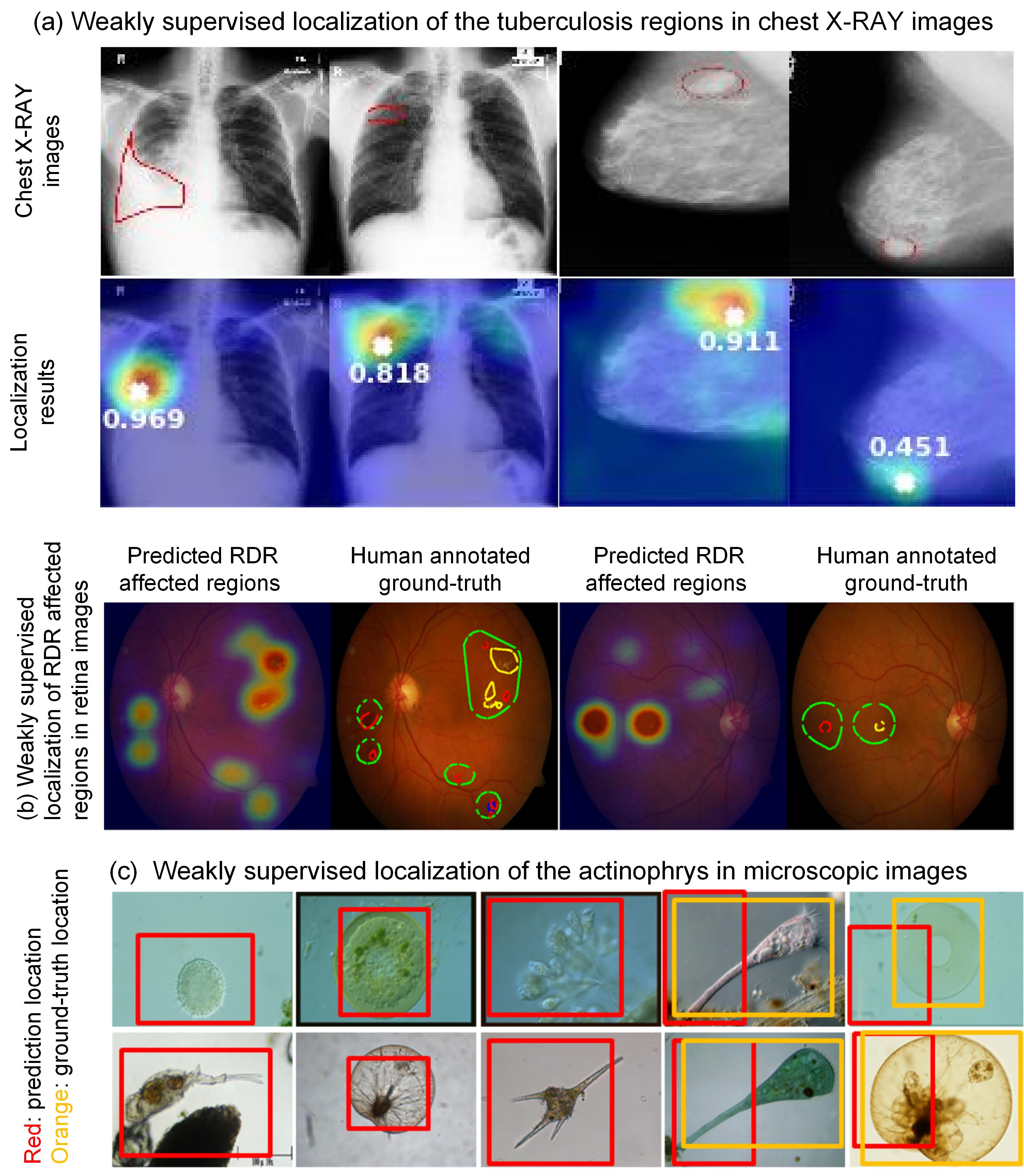}
 \caption{Examples of the application of weakly supervised object localization or detection approaches in medical image analysis. The examples are from \cite{hwang2016self,gondal2017weakly,li2015environmental}. }
 \label{bio}
 %\vspace{-0.6cm}
\end{figure}

As shown in Fig. \ref{bio}, medical image analysis is one area where the weakly supervised object localization and detection methods are of critical importance as only few annotations of target objects by trained experts in bio-image (e.g., organ or tissues).
To alleviate this problem, Hwang and Kim \cite{hwang2016self} develop a two-stream DNN model to localize the tuberculosis regions from the chest X-ray images.
%
%MH: useless and uninformative statement. What is the purpose of any training process?
%They propose to infer learning weights for the classification stream and localization stream during the training procedure.
 {Considering that the medical image-based applications usually do not have the pre-trained networks, this work proposes a weakly supervised learning scheme without requiring any pre-trained network parameters. The proposed network contains a fully connected layer-based classification branch and a CAM-based localization branch with shared convolutional layers for feature extraction. Both of the two branches are supervised by image label annotation, where a weighting parameter is introduced to dynamically control the relative importance between them to gradually switch the focus of the learning process from the classification branch to the localization branch. The authors demonstrate that the features learned from the classification layer at the early stage can provide informative cues to learn the localization branch at the late stage.}
%
%For detecting the specific types of retinal lesions, Wang et al. \cite{wang2018weakly} propose to model a fundus image as a superposition of background, blood vessels, and background noise.
%
%MH: BAD sentence to use "background" for two different things.
%They encode both the background knowledge of fundus images and the background noise into one unique weakly supervised learning model and corporately optimize the model using normal and abnormal images.
 {For detecting a general type of lesions, Wang et al. \cite{wang2018weakly} model the normal image as the combination of background and noise, while modeling the abnormal images as the combination of background, blood vessels, and noise. With the assumption that the noise for the normal image and abnormal image is the unified distribution, the image data can then be decomposed by the low-rank subspace learning technique to obtain the vessel areas. }
%
%MH: What unique weakly supervised learning? This is where is it important. Explain it clearly with some insight. Do not simply give long useless one/two sentence descriptions of a method. A laundry list of methods is useless!
%Both prior knowledge of fundus images and background noise into one unique weakly supervised learning model and corporately optimize the model using normal and abnormal images.
%
In \cite{gondal2017weakly}, Gondal et al. apply the weakly supervised object detection network on the retina images and achieve few false positives with high sensitivity on the lesion-level prediction.
Li et al. \cite{li2015environmental} apply a sparse coding-based weakly supervised learning method for localizing actinophrys in microscopic images.
Dubost et al. \cite{dubost2019weakly} propose weakly supervised regression neural networks for detecting brain lesions. Besides, some recent works also show great research interests in weakly supervised learning-based brain image analysis, such as brain disease prognosis \cite{liu2019weakly}, brain tumor or lesion segmentation \cite{ji2019scribble,wu2019weakly}, brain structure estimation \cite{bontempi2020cerebrum}, etc.
 {Notice that compared to common images, medical imaging data usually suffers from issues of low contrast and limited texture. Fortunately, some spatial priors could be obtained for different organs or lesions. These priors can be used to guide the weakly supervised learning process on medical imaging data.}

%MH: Did you have permision of authors to use their images?
%DW: Yes, have asked.

\subsection{Remote Sensing Imagery Analysis}
\begin{figure}[t]
 \centering
 \includegraphics[width=9.5 cm]{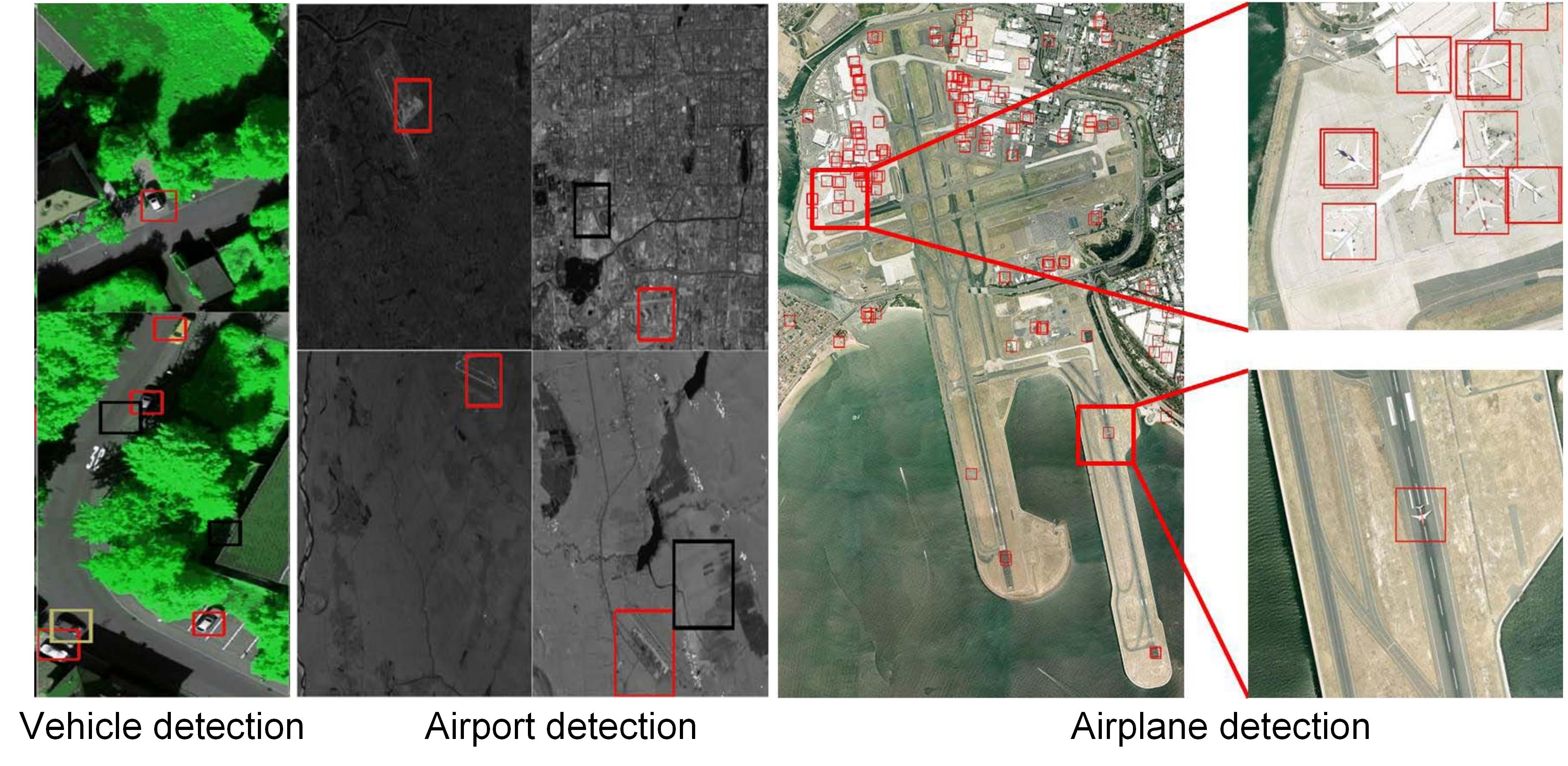}
 \caption{Examples of the application of weakly supervised object localization or detection approaches in remote sensing imagery analysis. The examples are from \cite{han2015object,zhang2016weakly}. }
 \label{remote}
 %\vspace{-0.6cm}
\end{figure}

Remote sensing imagery analysis is one of the most widely studied applications based on weakly supervised object localization and detection, where the input images are usually of large scale and the annotation process tends to be very time-consuming \cite{yao2020automatic,feng2020progressive,9069411,feng2020tcanet}(see Fig. \ref{remote}).
Zhang et al. \cite{Zhang2014weakly} propose a saliency-based weakly supervised detector learning method to learn the detectors of the airplane, vehicle, and airport from the remote sensing images collected from different sensors.
In this work, a weakly supervised object detection benchmark dataset for remote sensing imagery analysis is developed.
Han et al. \cite{han2015object} and Zhou et al. \cite{zhou2015negative} introduce the Bayesian inference and negative bootstrapping methods, respectively, for effective weakly supervised detectors for remote sensing images.
In \cite{zhang2016weakly}, Zhang et al. propose a coupled CNN method which combines a candidate region proposal network and a localization network to detect aircrafts in images.
 {When applying to remote sensing imagery analysis, the target objects are usually very small in size, which would dramatically increase the localization and detection difficulty given only weak supervision. }

%MH: useless paragraphs
%\subsection{Industry Environment}
%Weakly supervised object localization and detection methods have been developed in industry environmenst.
%
%For example, Zadrija et al. \cite{zadrija2018sparse} propose a Fisher-embedding-based weakly supervised object localization approach and apply it in on real traffic scenes to localize the zebra crossings and triangular warning signs. Vardazaryan et al. \cite{vardazaryan2018weakly} build a fully convolutional neural network-based end-to-end weakly supervised learning framework to localize surgical tools in endoscopic videos. Kanezaki et al. \cite{kanezaki2013weakly} propose a 3D feature-based weakly-supervised learning method for object detection on the color and depth images captured by a Kinect sensor in the real environment. Mehta et al. \cite{mehta2018deepsolareye} propose a weakly supervised deep neural network learning approach with a bi-directional input-aware fusion block to localize soil regions from the solar panel images.

%\subsection{Application in Astronomical-image Analysis}

%\section{Discussions on Learning Stratifies}

\section{Future Directions}

%Although there have been numerous methods in this field, the performance of the existing approaches are still not satisfactory and many problems in this field still remain unsolved. Thus, in this section, we point out several potential future research direction for developing stronger weakly supervised object localization and detection systems.
We discuss the issues to be addressed in this field for future research.

%MH: bad and useless argument... anyone can do that. why is it important? why is imporant to discuss it here?
%\subsection{Stronger Network Backbones}

%One of the most direct ways to improve the performance of the current weakly supervised object localization and detection approaches is to build the weakly supervised learning framework based on stronger network backbones. As we know, most of the currency weakly supervised object localization and detection approaches are based on the Fast RCNN \cite{girshick2015fast} or some simpler image classification network backbones, such as the AlexNet \cite{krizhevsky2012imagenet} and VGGnet \cite{simonyan2014very}. However, with the rapid development of the deep learning technique in the conventional fully supervised learning domain, there have already been a serious of newer and stronger network backbones, such as YOLO \cite{redmon2017yolo9000}, SSD \cite{liu2016ssd}, FPN \cite{lin2017feature}, SNIP \cite{singh2018analysis} for object detection and ResNet \cite{he2016deep}, DenseNet \cite{huang2018condensenet}, IGCV \cite{xie2018igcv} for classification, that can be used to improve the weakly supervised object localizer and detector learning process. The researchers are encouraged to integrate the existing weakly supervised learning network modules and strategies or the novel ones with the stronger network backbones for better learning performance.

\subsection{Multiple Instance Learning}
%Besides the network backbone, the MIL model is another key factor to improve the weakly supervised learning performance.
%
Weakly supervised object localization or detection methods can be easily formulated within the MIL framework.
Early methods in this filed usually add prior knowledge \cite{song2014learning} or post regularization \cite{bilen2014weakly} on the classic MIL models, such as LSVM \cite{yu2009learning}, while the current research obtains the breakthrough by building deep MIL models \cite{wang2018revisiting,zhou2016learning,wu2015deep}. To further improve the weakly supervised learning performance, efforts should be made to introduce the most advanced ideas and techniques in the research filed of MIL, such as the set-level problem \cite{xu2019isolation} the key instance shift issue \cite{zhang2017multi} and the scalable issue \cite{huang2018fast} in MIL.
Further research towards more advanced MIL techniques would also bring helpful insights for the WSOL and WSOD in the future.

\subsection{Multi-Task Learning}
Another future direction is to combine multiple weakly supervised learning tasks into a unified learning framework. These tasks may include object detection \cite{girshick2015fast}, semantic segmentation \cite{Chen2018DeepLab}, instance segmentation \cite{khoreva2017simple}, 3D shape reconstruction \cite{Fan2017A}, and depth estimation \cite{Godard_2017_CVPR}.
Essentially, efforts for simultaneously accomplishing multiple aforementioned tasks have been made in the conventional fully supervised learning scenario \cite{khoreva2017simple,hariharan2014simultaneous,zhao2005closely,wu2007simultaneous,zhu2015segdeepm}, which have demonstrated that such learning mechanism can bring helpful information from one task to the other ones.
The methods proposed by Zhang et al. \cite{zhang2017learning} is an early attempt to implement such a weakly supervised multi-task learning mechanism and the experimental results show that training object segmentation and 3D shape reconsecration models jointly indeed benefits the both weakly supervised learning tasks. With similar spirits to \cite{zhang2017learning}, Zhang et al. \cite{zhang2018spftn} and Shen et al. \cite{shen2019cyclic} establish a self-paced fine-tuning network and a cyclic guidance network to jointly learn object localization and segmentation models under the weak supervision, respectively. {Under the multi-task weakly supervised learning scenario, one key problem is that the learning ambiguity of each individual task might be aggregated and the imprecise prediction on one task might affect the learning on other tasks. To deal with this problem, one needs to disentangle the complex multi-task learning, separately learning each individual task first, and then leveraging the confidence knowledge from each task to provide informative priors to guide the learning processes of the other tasks. }

%\begin{figure}[t]
% \centering
% \includegraphics[width=9.5 cm]{indapp.jpg}
% \caption{Examples of the application of weakly supervised object localization or detection approaches in industry environment. The examples are from \cite{inoue2018cross,gonthier2018weakly,crowley2013gods}, where detection results in different colors in the solar panel images and the road images indicate different types of objects. }
% %\vspace{-0.6cm}
%\end{figure}

\subsection{Robust Learning Theory}
To address the learning under uncertainty issue that is inherently existed in the weakly supervised learning process, robust learning strategy will become one of the key techniques in the future. The goal is to alleviate the influence of the noisy samples during the learning process. In implementation, such learning strategy is usually achieved by selecting easy and confident training samples in the early learning stages while using hard and more ambiguous training samples in the late learning stages. Essentially, a number of recent methods have already introduced the robust learning strategies into their learning frameworks. For example, Shi and Ferrari \cite{shi2016weakly} propose a curriculum learning strategy to feed training images into the WSOL learning loop in order from images containing bigger objects down to smaller ones. The training order is determined by the size of the object, which is estimated based on a regression model. Similarly, Zhang et al. \cite{zhang2018zigzag} design a zigzag learning strategy, where they first develop a criterion to automatically rank the localization difficulty of an image, and then learn the detector progressively by feeding examples with increasing difficulty. As can be seen, these methods are just intuitive ways to introduce robust learning strategy into the weakly supervised object localization and detection frameworks, while they have already achieved obvious performance gains when compared with the conventional learning strategy. Along this line, Zhang et al. \cite{Zhang2018Leveraging} propose a self-paced curriculum learning framework for weakly supervised object detection. By integrating the curriculum learning \cite{bengio2009curriculum} with the self-paced learning \cite{kumar2010self}, the established learning framework provides a more theoretical-sounded way to improve the learning robustness. However, the solid robust learning theory is still lack in this research field.

\subsection{Reinforcement and Adversarial Learning}
Besides the conventional CNN models, it is also worth trying to apply some more advanced learning models into the learning process of the weakly supervised object detector. Here we give two examples. The first one is the deep reinforcement learning.
According to~\cite{larochelle2010learning}, biological vision systems are believed to have a sequential process with changing retinal fixations that gradually accumulate evidence of certainty when searching or localizing objects. Several existing methods~\cite{caicedo2015active,jie2016tree,liang2017deep,huang2017learning, yoo2017action} have also demonstrated that designing deep reinforcement learning frameworks to model such a sequential searching process can indeed help to address the object localization, detection, and tracking problems in the computer vision community. Thus, it is highly desirable, both biologically and computationally, to explore deep reinforcement learning models that facilitate the weakly supervised object localization and detection systems in such a sequential searching process \cite{zhang2020discriminant}.
The second one is the generative adversary learning. As we know, generative adversary learning has been demonstrated to have advantages in unsupervised and semi-supervised learning scenarios \cite{goodfellow2014generative,tsai2018learning,springenberg2015unsupervised,shrivastava2017learning}. It can generate the desired data distribution based on very weak supervision, i.e, ``real'' or ``fake''. Such capacity endows generative adversary learning very large potential in solving the weakly supervised object localization and detection problems. Although existing methods, such as \cite{shen2018generative,zhang2018adversarial,diba2019weakly}, have already made efforts to introduce such a learning mechanism into the weakly supervised object localization and detection, there is still much room for improvement along this research direction.

\subsection{Prior-guided Deep MIL}
From Table \ref{table endtoend} and Table \ref{table multinet}, we can observe that most of the current deep weakly supervised object detection methods have not introduced any prior knowledge into their learning frameworks. However, from our review on classic models (see Sec. \ref{Classic Models}), prior knowledges actually play important roles in avoiding the weakly supervised learning process from drifting to trivial solutions. Considering this issue, some recent works utilize prior knowledges of saliency \cite{li2018weakly}, objectness \cite{rahimipairwise,zhong2020boosting}, shape \cite{li2017multiple}, count \cite{gao2018c,Learning2020}, human action \cite{yang2019activity}, human object interaction \cite{kim2019tell}, mask-out scoring \cite{xu2019adaptively} in their frameworks. However, research towards building effective deep MIL frameworks (such as the one with prior knowledge distillation \cite{chen2017learning} or cross domain adaptation \cite{Hong_2018_CVPR}) to embed helpful prior knowledge into the weakly supervised learning process needs to be further explored in the future. In addition, the co-occurring patterns mined in co-saliency detection \cite{zhang2016co,zhang2016detection} and object co-localization \cite{shaban2019learning,wei2019unsupervised} approaches can also be used as informative priors to guide the deep multiple instance learning process in weakly supervised object localization and detection.

\section{Conclusions}
In this paper, we provide a comprehensive survey of existing literatures in the research field of weakly supervised object localization and detection. We start with the introduction of the definition of the task and the key challenges that make the weakly supervised learning process hard to implement. Then, we introduce the development history of this field, the taxonomy of methods for weakly supervised object localization and detection, and the relationship between different categories. After reviewing existing literatures in each category of methodology, we introduce the benchmark datasets and evaluation metrics that are widely used in this filed, which are followed by the reviewing of the applications of the existing weakly supervised object localization and detection algorithms. Finally, we point out several future directions that may further promote the development of this research field.

{\small
\bibliographystyle{ieee}
\bibliography{WSsurvey}
}


\begin{IEEEbiography}[{\includegraphics[width=1in,height=1.25in,clip,keepaspectratio]{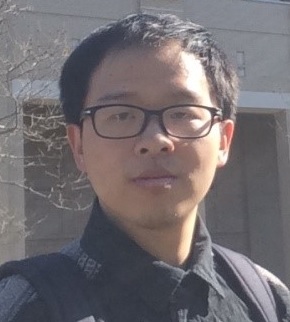}}]{Dingwen Zhang}
received his Ph.D. degree from the Northwestern Polytechnical University, Xi'an, China, in 2018. He is currently a full professor in School of Automation, Northwestern Polytechnical University. From 2015 to 2017, he was a visiting scholar at the Robotic Institute, Carnegie Mellon University. His research interests include computer vision and multimedia processing, especially on saliency detection, video object segmentation, and weakly supervised learning.
\vspace*{-30pt}
\end{IEEEbiography}

\begin{IEEEbiography}[{\includegraphics[width=1in,height=1.25in,clip,keepaspectratio]{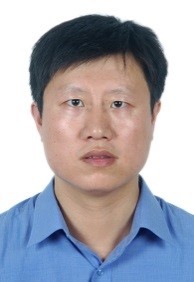}}]{Junwei Han}
(M'12-SM'15) is a Professor with Northwestern Polytechnical University, Xi'an, China. He received Ph.D. degree in Northwestern Polytechnical University in 2003. He was a Research Fellow in Nanyang Technological University, The Chinese University of Hong Kong, and University of Dundee. His research interests include computer vision and brain imaging analysis. He has published over 100 papers in IEEE TRANSACTIONS and top tier conferences. He is currently an Associate Editor of IEEE Trans. on Neural Networks and Learning Systems, IEEE Trans. on Circuits and Systems for Video Technology, IEEE Trans. Cognitive and Developmental Systems, IEEE Trans. on Human-Machine Systems, Neurocomputing, and Machine Vision and Applications.
\vspace*{-30pt}
\end{IEEEbiography}

\begin{IEEEbiography}[{\includegraphics[width=1in,height=1.25in,clip,keepaspectratio]{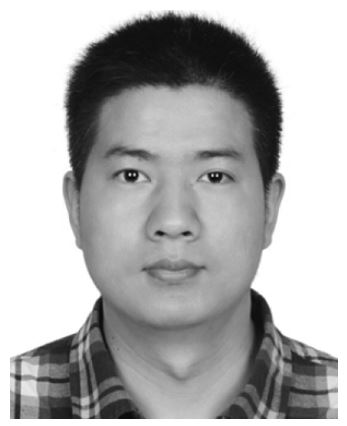}}]{Gong Cheng}
received the B.S. degree from Xidian University, Xi'an, China, in 2007, and the M.S. and Ph.D. degrees from Northwestern Polytechnical University, Xi'an, in 2010 and 2013, respectively. He is currently a Professor with Northwestern Polytechnical University. His main research interests are computer vision and pattern recognition.
\vspace*{-30pt}
\end{IEEEbiography}

\begin{IEEEbiography}[{\includegraphics[width=1in,height=1.25in,clip,keepaspectratio]{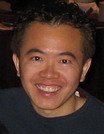}}]{Ming-Hsuan Yang}
received the PhD degree in computer science from the University of Illinois at Urbana-Champaign, in 2000. He is a professor in Electrical Engineering and Computer Science from the University of California, Merced. He served as an associate editor of the IEEE Transactions on Pattern Analysis and Machine Intelligence from 2007 to 2011, and is an associate editor of the International Journal of Computer Vision, the Computer Vision and Image Understanding, the Image and Vision Computing, and the Journal of Artificial Intelligence Research. He received the NSF CAREER award in 2012, and the Google Faculty Award in 2009. He is a Fellow of the IEEE and Senior Member of the ACM.
\vspace*{-30pt}
\end{IEEEbiography}


%\vfill


\end{document}